\PassOptionsToPackage{hyphens}{url} % allow URL line-break at hyphens (must precede hyperref load)
\documentclass{article}

\usepackage[preprint]{corl_2026} % arXiv preprint: shows authors, removes CoRL footnote, no line numbers
\usepackage{booktabs}
\usepackage{amsmath,amssymb}
\usepackage{graphicx}
\usepackage{subcaption}
\usepackage{multirow}
\usepackage{makecell}
\usepackage{xcolor}
\usepackage{colortbl}
\usepackage{microtype}
\usepackage{xspace}
\usepackage{wrapfig}
\usepackage{pifont}
\usepackage{placeins} % \FloatBarrier to flush figures/tables before section breaks

\newcommand{\cmark}{\ding{51}}
\newcommand{\xmark}{\ding{55}}
\newcommand{\blfootnote}[1]{%
  \begingroup\renewcommand\thefootnote{}\footnote{#1}\addtocounter{footnote}{-1}\endgroup}

\definecolor{skipblue}{rgb}{0.800,0.941,1.000}
\definecolor{mydarkblue}{rgb}{0,0.08,0.45}

\title{\textsc{SKIP}: Sparse Keyframe Interpolation Paradigm\\
for Efficient Embodied World Models}

\author{%
\begin{minipage}{0.95\linewidth}\centering\normalfont
{\bfseries Ziheng He\textsuperscript{1,2}, Yixiang Chen\textsuperscript{1,2}, Ning Yang\textsuperscript{1,3}, Zhanqian Wu\textsuperscript{4}, Qisen Ma\textsuperscript{1,2}, Yuan Xu\textsuperscript{1,2}, Jiabing Yang\textsuperscript{1,2}, Peiyan Li\textsuperscript{1,2}, Xiangnan Wu\textsuperscript{1,2}, Xiaofeng Wang\textsuperscript{4,5}, Zheng Zhu\textsuperscript{4,5}, Jing Liu\textsuperscript{6}, Nianfeng Liu\textsuperscript{6}, Yan Huang\textsuperscript{1,2,6,*}}\\[6pt]
{\small
\textsuperscript{1}UCAS \quad \textsuperscript{2}CASIA \quad \textsuperscript{3}NJU \quad \textsuperscript{4}GigaAI \quad \textsuperscript{5}THU \quad \textsuperscript{6}FiveAges\\
\texttt{heziheng261@mails.ucas.ac.cn}}
\end{minipage}%
}

\begin{document}
\maketitle
\blfootnote{\textsuperscript{*}\,Corresponding author.}

%===============================================================================
\begin{abstract}
Embodied world models have emerged as a promising paradigm in robotics by predicting how robot actions affect the surrounding scene.
However, the rollout inference remains computationally expensive in pixel space, as long-horizon manipulation videos typically have to be generated frame by frame.
This cost cannot be easily reduced by indiscriminately dropping frames, since downstream policies rely on complete preservation of sparse task-relevant events such as approach, contact, grasp, and release.
To address this challenge, we propose Sparse Keyframe Interpolation Paradigm (\textsc{SKIP}), an event-preserving sparse-to-dense framework that avoids dense frame-by-frame generation.
\textsc{SKIP} first identifies task-relevant keyframes by leveraging robot-aware multimodal features.
It then synthesizes only these keyframes with a sparse video diffusion model. A learned gap predictor and an action-conditioned interpolator subsequently reconstruct the missing intervals according to the robot actions.
On LIBERO, \textsc{SKIP} generates dense rollouts $4.16\times$ faster than a dense baseline while improving visual fidelity and reducing aggregate FVD by $89.0\%$.
Importantly, \textsc{SKIP}-generated videos are effective policy-training data. Even when they fully replace real demonstrations, $\pi_{0.5}$ success drops only $1.3$ pp in LIBERO simulation and $6.7$ pp on the real robot, whereas fully dense frame-by-frame generation collapses by $48$ to $58$ pp.
\end{abstract}


%===============================================================================
\section{Introduction}
\label{sec:intro}

\begin{figure}[!htbp]
  \centering
  \includegraphics[width=1.0\linewidth]{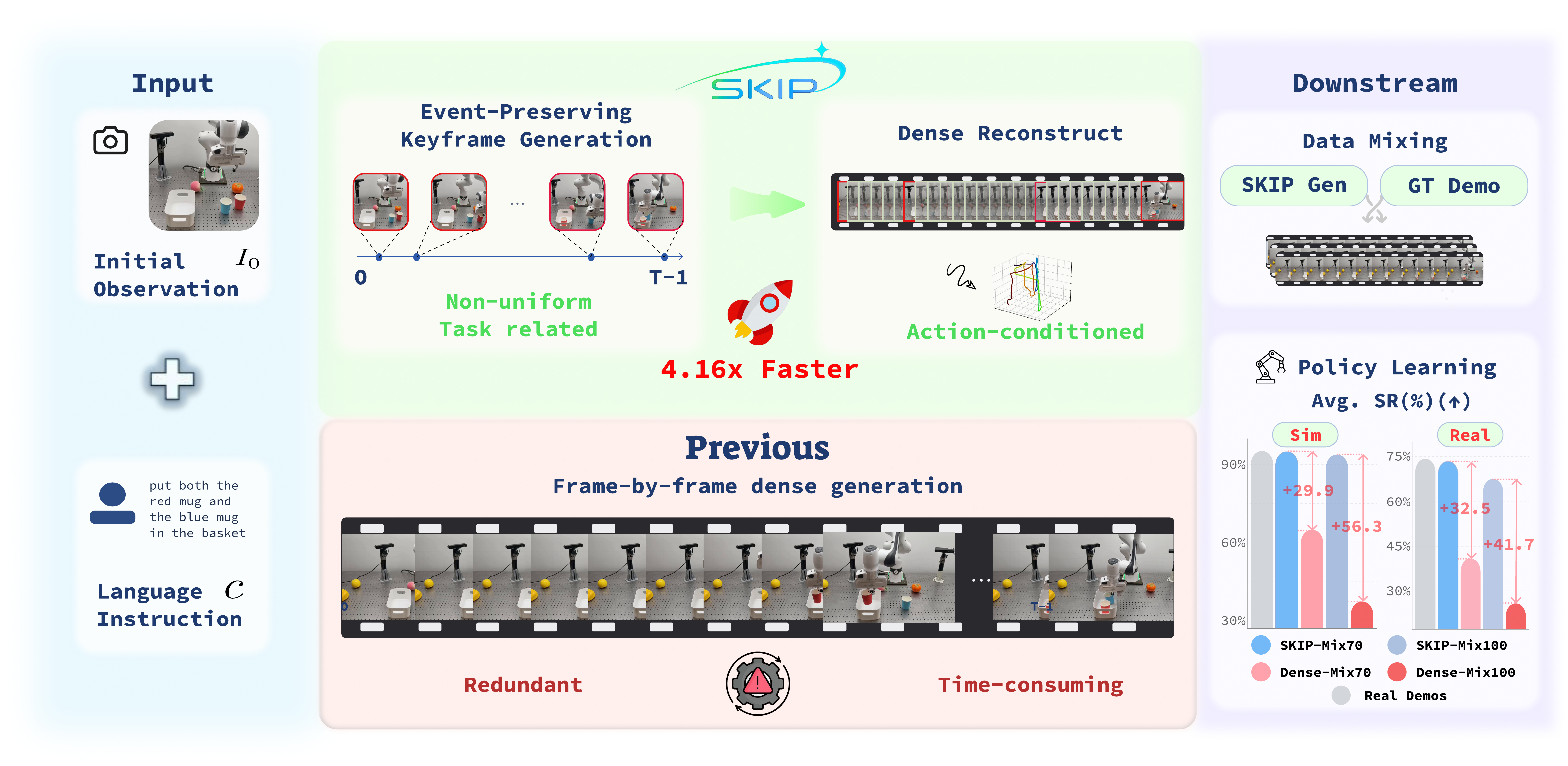}
  \caption{\textbf{Overview of \textsc{SKIP}.} Our framework predicts event-preserving sparse keyframes from an initial observation and a language instruction, then recovers the dense rollout via learned gap prediction and action-conditioned interpolation.}
  \label{fig:pipeline}
\end{figure}

Recent Vision-Language-Action (VLA) models have achieved rapid progress in robot manipulation~\citep{brohan2023rt,zitkovich2023rt,kim2025openvla,team2024octo,black2024pi_0,black2025pi_,li2026bridgevla}.
However, their performance still depends heavily on large-scale robot demonstrations for training and on real-world trials for behavior validation.
As both are costly and time-consuming, there is strong motivation to develop embodied world models capable of predicting how candidate robot actions unfold in a given scene. Such predictions can both supply synthetic training data and verify behavior without real-world execution.
In particular, generative video world models make these predictions explicit as pixel-level rollouts.

However, this explicitness usually requires high inference cost, because current generative video world models~\citep{peebles2023scalable,videoworldsimulators2024,yang2025cogvideox,wan2025wan} typically implement diffusion computation frame by frame. It is redundant since most neighboring frames contain similar backgrounds and manipulator-motion patterns.
A straightforward alternative is temporal sparsification, explored in video summarization~\citep{gygli2014creating,zhou2018deep}, temporal action localization~\citep{lin2019bmn,zhang2022actionformer}, and hierarchical video generation~\citep{zhao2025moviedreamer,wang2025keyvid}.
However, the criteria for selecting keyframes in these areas are quite different from what embodied manipulation requires.

For example, in a fixed-camera manipulation rollout, where usually only the robot arm and the manipulated object move, the task-relevant information concentrates at a few discrete events such as an approach end, a gripper closing, a contact, or a release. These events are separated by smooth free-space motion that is visually near-redundant and similar in motion patterns, which is exactly where uniform sampling and generic-video criteria become misaligned.
The difficulty of interpolation also varies significantly across frames. Dropping even a few frames during a gripper closure can miss the critical moment that determines whether the rollout successfully grasps the object. These frames often exhibit only a few pixels of visual change, making their loss abrupt and unrecoverable. Preserving these events is therefore essential, yet generic sparse-video methods ignore this embodiment-specific structure and routinely drop them.

To address this gap, we propose \textsc{SKIP}, an event-preserving sparse-to-dense framework for efficient explicit rollout inference in generative video world models.
As illustrated in Figure~\ref{fig:pipeline}, \textsc{SKIP} decouples dense rollout synthesis into two phases, event-preserving keyframe generation and dense reconstruction, realized by three modules. \textsc{SKIP-Selector} identifies event-preserving keyframes through robot-aware multimodal fusion, which supervise \textsc{SKIP-Generator} to synthesize only these keyframes. \textsc{SKIP-Reconstructor} then recovers the dense rollout with a learned gap predictor and an action-conditioned interpolator. At inference, only \textsc{SKIP-Generator} and \textsc{SKIP-Reconstructor} run, realizing the two phases and producing dense action-aligned rollouts ready for downstream policy training and behavior-level inspection.

Our contributions are summarized as follows. \textbf{(1)} We show that a few decisive manipulation events can largely replace dense frame-by-frame generation, supporting both policy training and behavior preview. \textbf{(2)} We realize this paradigm as \textsc{SKIP}, in which \textsc{SKIP-Selector} selects event-preserving keyframes through robot-aware multimodal fusion, \textsc{SKIP-Generator} synthesizes only these keyframes, and \textsc{SKIP-Reconstructor} recovers the dense action-aligned rollout via learned gap prediction and action-conditioned interpolation. \textbf{(3)} On LIBERO~\citep{liu2023libero}, \textsc{SKIP} runs $4.16\times$ faster than a dense baseline with higher fidelity. Moreover, even fully replacing real demonstrations for $\pi_{0.5}$~\citep{black2025pi_} training drops success by only $1.3$ pp in LIBERO simulation and $6.7$ pp on a four-task Franka real robot, versus a $48$ to $58$ pp collapse for fully dense-generated rollouts.

%===============================================================================
\section{Related Work}
\label{sec:rw}

\textbf{Embodied world models.}
Embodied world models~(EWMs) predict how candidate robot actions unfold in the environment, supporting policy learning, planning, and behavior inspection.
Latent or embedding-space EWMs such as Dreamer~\citep{hafnerdream,hafner2025mastering}, TD-MPC2~\citep{hansen2024td}, and V-JEPA~\citep{bardesrevisiting,assran2025v} roll out in compact internal states.
These are efficient for control but inaccessible for pixel-level inspection or VLA training.
Generative video EWMs such as IRASim~\citep{zhu2025irasim} synthesize explicit pixel rollouts directly usable for policy learning and inspection, but their dense frame-by-frame generation makes inference time-consuming for long manipulation trajectories.
A related line of work scales robot data without new teleoperation through simulation, replay, or augmentation~\citep{mandlekar2023mimicgen,chen2023genaug,wang2024robogen}, reshaping how data is constructed but leaving the per-rollout generative cost untouched.
\textsc{SKIP} instead targets this per-rollout cost, decoupling sparse keyframe synthesis from lightweight action-guided recovery to keep pixel-level rollouts while removing redundant frame-by-frame computation.

\textbf{Keyframe extraction and event preservation.}
Keyframe extraction is widely used to reduce temporal redundancy in video summarization~\citep{gygli2014creating,zhou2018deep}, temporal action localization~\citep{lin2019bmn,zhang2022actionformer}, and hierarchical video generation~\citep{zhao2025moviedreamer,wang2025keyvid}, but these criteria are misaligned with manipulation's event-driven structure.
The underlying sparse-to-dense principle is also effective in adjacent generative domains such as human motion diffusion~\citep{bae2025less}, but has not been adapted to action-conditioned embodied video world models.
The closest robot-specific efforts still select keyframes at a coarse, single-modality level. KeyWorld~\citep{li2025keyworld} uses pose-only RDP simplification, and RoboEnvision~\citep{yang2025roboenvision} uses language-defined subtask boundaries with visual-only interpolation.
\textsc{SKIP} instead selects fine-grained physical events through robot-aware multimodal fusion and validates the generated keyframe videos as a substitute for real demonstrations in both simulation and on a real robot.

\textbf{Action-conditioned dense reconstruction.}
Recovering continuous video from sparse frames is a standard generative task, but embodied AI requires the recovered intermediates to align with robot actions.
Generic frame interpolation~\citep{jiang2018super,huang2022real,li2023amt} is efficient but action-agnostic, failing to model variable pauses or gripper-state changes when relying on visual endpoints alone.
Action-conditional video prediction~\citep{oh2015action,finn2017deep,ebert2018visual} captures action dynamics but defaults to dense, frame-by-frame generation.
\textsc{SKIP} addresses this trade-off with a lightweight action-guided interpolator that fuses visual anchors with continuous action sequences to maintain tight action alignment without the dense diffusion overhead.

%===============================================================================
\section{\textsc{SKIP}}
\label{sec:method}

\subsection{Problem formulation and overview}
\label{sec:setup}

Our objective is an efficient video world model that simulates how candidate robot actions unfold in the physical world, using a dataset of expert demonstrations for downstream planning, behavior validation, and policy training.
Each demonstration $\tau=(\mathcal{I},\mathcal{A},\mathcal{S},c)$ consists of an RGB observation sequence $\mathcal{I}=\{I_t\}_{t=0}^{T-1}$, an expert action sequence $\mathcal{A}=\{\mathbf{a}_t\}_{t=0}^{T-2}$, an optional proprioceptive stream $\mathcal{S}=\{\mathbf{s}_t\}_{t=0}^{T-1}$, and a language instruction $c$.
Given the initial observation $I_0$, the language instruction $c$, and the future actions $\mathcal{A}$, our model predicts the future RGB observations $\hat{\mathcal{I}}_{1:T-1}$, where $\hat{\cdot}$ marks model-generated quantities throughout.

\textsc{SKIP} consists of three modules (Figure~\ref{fig:method_detail}): \textsc{SKIP-Selector}, \textsc{SKIP-Generator}, and \textsc{SKIP-Reconstructor}, detailed in the following subsections.

\subsection{\textsc{SKIP-Selector} and \textsc{SKIP-Generator}: Event-preserving keyframe generation}
\label{sec:kts}

\begin{figure}[!htbp]
  \centering
  \includegraphics[width=1.0\linewidth]{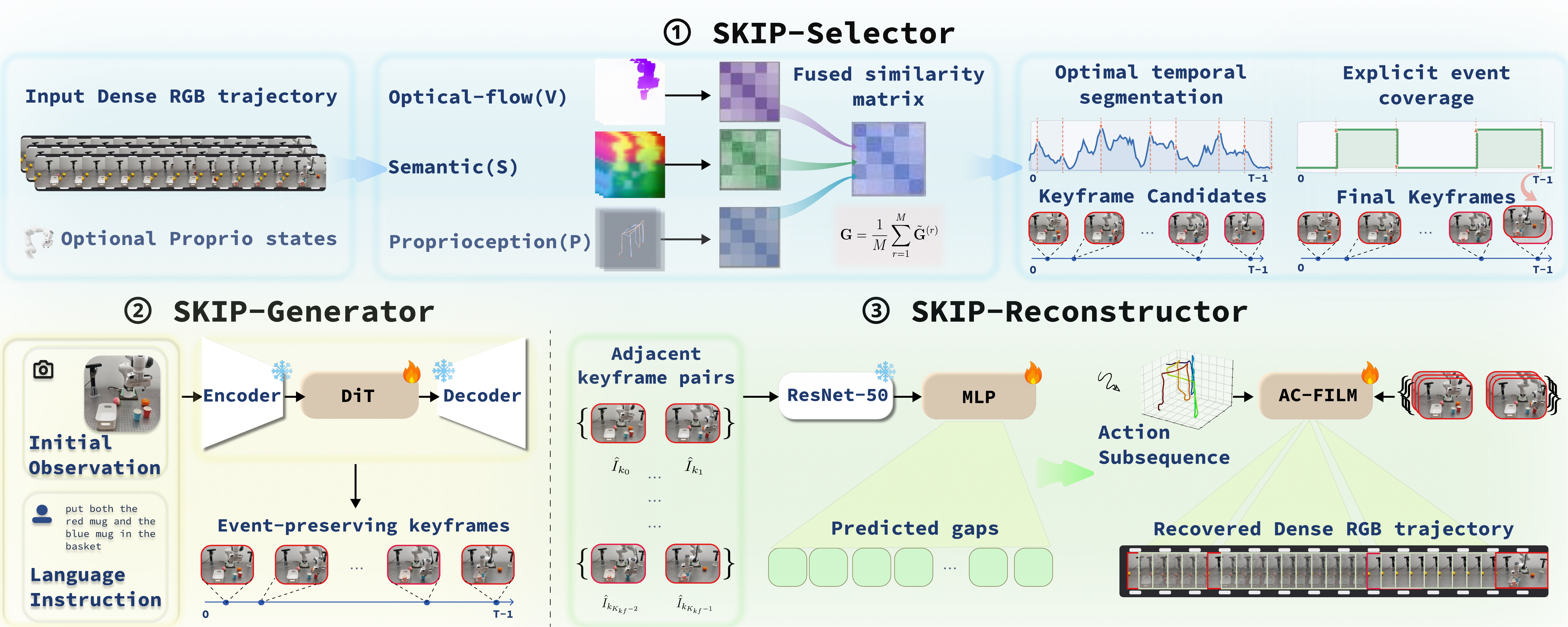}
  \caption{\textbf{SKIP architecture.}
  \textbf{\textsc{SKIP-Selector}:} fuses visual, semantic, and optional proprioceptive features into similarity matrices, applies temporal segmentation, and enforces gripper-event coverage to yield a sparse set of event-preserving keyframes.
  \textbf{\textsc{SKIP-Generator}:} a fine-tuned video diffusion model synthesizes only these keyframes from the initial observation and language instruction.
  \textbf{\textsc{SKIP-Reconstructor}:} at synthesis time a gap predictor estimates the temporal gap between successive generated keyframes, and an action-conditioned interpolator runs over the corresponding action subsequences to reconstruct a dense rollout for policy training.}
  \label{fig:method_detail}
\end{figure}

Keyframe selection criteria designed for generic video tasks are not well-suited to embodied manipulation.
Content-agnostic and feature-clustering heuristics ignore the event-driven temporal structure of manipulation rollouts, while the closest robot-specific criterion, pose-only geometric simplification~\citep{li2025keyworld}, depends on full proprioception that is often unavailable in video-only datasets and overlooks visual events with a small pose signature.

Motivated by these task-specific demands, we introduce \textsc{SKIP-Selector}.
As depicted in the top panel of Figure~\ref{fig:method_detail}, \textsc{SKIP-Selector} processes the dense RGB trajectory alongside an optional proprioceptive stream to output a sparse keyframe index set $\mathcal{K}$ of user-configurable size $K_{\mathrm{kf}}$.
Each frame is represented by three complementary streams: optical-flow features~(V) for dense manipulator motion, frozen DINO features~(S)~\citep{simeoni2025dinov3,oquab2024dinov2} for semantic scene changes, and optional proprioceptive features~(P) for contact events invisible to vision.

\textbf{Robot-aware similarity fusion.}
We fuse these heterogeneous streams into a unified similarity representation.
Direct feature concatenation would allow high-dimensional streams to dominate low-dimensional signals in the distance metric.
To address this imbalance, we compute an independent RBF kernel matrix~\citep{scholkopf2005learning} for each available modality $r$:
$G^{(r)}_{ij}{=}\exp(-\|\mathbf{f}^{(r)}_i{-}\mathbf{f}^{(r)}_j\|^2/(2\sigma_r^2))$ measures the similarity between frames $i$ and $j$, with $\sigma_r$ set to the median pairwise distance.
We then center and trace-normalize each modality matrix before averaging them into a fused similarity matrix $\mathbf{G}=\frac{1}{M}\sum_{r=1}^{M} \tilde{\mathbf{G}}^{(r)}$, where $M$ is the number of available modalities.
This fusion constrains each modality within its own geometry and allows all feature streams to contribute comparably.

\textbf{Optimal temporal segmentation.}
We next segment the fused similarity matrix $\mathbf{G}$ into contiguous task stages using Kernel Temporal Segmentation (KTS)~\citep{potapov2014category}, which partitions an ordered sequence into temporal segments rather than clustering frames independently, and controls segment count through a penalty without requiring an event classifier or pre-specified event types.
KTS applies dynamic programming to identify boundaries that minimize within-segment variance:
\begin{equation}
  \min_{L,\{b_j\}} \sum_{j=0}^{L-1}\!\left[\sum_{i=b_j}^{b_{j+1}-1}\!G_{ii}-\frac{1}{b_{j+1}-b_j}\!\sum_{i,i'\in[b_j,b_{j+1})}\!G_{ii'}\right]+\lambda L.
  \label{eq:kts}
\end{equation}
Since generative video backbones require fixed-length inputs, we expose $K_{\mathrm{kf}}$ as a user-configurable budget and bisection-search the penalty $\lambda$ until KTS yields exactly that many intervals.

\textbf{Explicit event coverage.}
KTS optimizes macroscopic stage boundaries, and we add a lightweight, optional post-processing step to explicitly protect short discrete events such as gripper transitions.
When gripper commands are available, we detect sign changes
$\mathcal{E}_g{=}\{t:\mathrm{sign}(g_t){\neq}\mathrm{sign}(g_{t-1})\}$ and enforce coverage by replacing the nearest non-event candidate keyframe with the missed event frame.
This post-processing preserves $|\mathcal{K}|=K_{\mathrm{kf}}$, requires no event classifier or extra annotation, and directly protects the events that matter for policy learning.

\textbf{Sparse keyframe generation.}
\textsc{SKIP-Selector} runs only at training time, where it selects the indices $\mathcal{K}$ from the dense ground-truth trajectory and turns the ordered sparse sequence $\mathcal{I}_{\mathcal{K}}=\{I_k:k\in\mathcal{K}\}$ into the supervision target.
We fine-tune Wan~2.2-TI2V-5B~\citep{wan2025wan} on this target conditioned only on the initial observation $I_0$ and language instruction $c$, giving the objective $\min_{\theta}\, \mathbb{E}_{\tau}\!\left[\mathcal{L}_{\mathrm{Wan}}\!\left(\theta;\,I_0,c,\mathcal{I}_{\mathcal{K}}\right)\right]$.
At inference the dense trajectory is unavailable and \textsc{SKIP-Selector} does not run, so the fine-tuned \textsc{SKIP-Generator} instead produces the ordered $K_{\mathrm{kf}}$ keyframes $\hat{\mathcal{I}}_{\mathcal{K}}$ directly from $(I_0,c)$.
\textsc{SKIP-Generator} is thus action-free: the future actions $\mathcal{A}$ enter only at the \textsc{SKIP-Reconstructor} stage (\S\ref{sec:recon}), and keyframe-action consistency holds because in our offline data-augmentation setting the learned keyframes and the paired actions come from the same expert trajectories.

\subsection{\textsc{SKIP-Reconstructor}: Gap prediction and action-conditioned recovery}
\label{sec:recon}

Using the sparse generated keyframes, we reconstruct a dense action-paired rollout in two steps. We first predict the number of intermediate frames in each inter-keyframe segment with a gap predictor, then fill them in with an action-conditioned flow interpolator, namely AC-FILM, guided by the corresponding action subsequence.

\textbf{Gap prediction.}
For a selected keyframe sequence $\mathcal{K}=\{k_i\}_{i=0}^{K_{\mathrm{kf}}-1}$, the ground-truth gap for segment $i$ is $g_i=k_{i+1}-k_i$. During training, we encode adjacent target keyframes with a frozen image encoder and regress $g_i$ from the concatenated features $[\mathbf{e}_i,\mathbf{e}_{i+1},\mathbf{e}_{i+1}-\mathbf{e}_i]$. At synthesis time, we apply the same predictor to adjacent generated keyframes, then normalize the predicted lengths and convert them to integers so that $\sum_i \hat{g}_i=T-1$, which preserves the original rollout length and decouples \textsc{SKIP-Generator} from absolute timing.

\textbf{Action-conditioned interpolation.}
Each inter-keyframe segment covers a short temporal interval, and the corresponding action subsequence specifies the underlying motion. Under this setting, a flow-based interpolator with low per-frame compute is sufficient as the backbone. We adopt FILM~\citep{reda2022film} for its stability under the large endpoint displacement between sparse keyframes. AC-FILM retains its recursive midpoint protocol, which subdivides each segment until reaching the target length. For segment $i$, we select the action sub-sequence $\mathbf{A}_i$ of length $\hat{g}_i$ and map it to a context vector $\mathbf{z}_i$ with a lightweight $1$D encoder using midpoint-query attention. We then use per-level FiLM projections~\citep{perez2018film} to convert $\mathbf{z}_i$ into modulation parameters $\boldsymbol{\gamma}_\ell$ and $\boldsymbol{\beta}_\ell$ that scale and shift each pyramid level as $\mathbf{F}_\ell'=\boldsymbol{\gamma}_\ell\odot\mathbf{F}_\ell+\boldsymbol{\beta}_\ell$. To suppress modulation when the robot is nearly static, we gate the modulation through a convex blend $\mathbf{F}_\ell''=s\,\mathbf{F}_\ell'+(1-s)\,\mathbf{F}_\ell$, where the action-magnitude score $s=\sigma(w_s\bar{m}+b_s)$ is derived from the mean pose-increment norm $\bar{m}$. As a result, the module is conservative on low-action intervals and action-aware around contact-rich motion. At inference, we apply AC-FILM recursively within each predicted gap to produce the dense action-paired rollout.
%===============================================================================
\section{Experiments}
\label{sec:exp}

We organize our experiments around three questions. \textbf{(i) Sparse-to-dense generation.} Does the proposed paradigm improve efficiency and generation quality over dense-generation and keyframe-selection baselines? \textbf{(ii) Replacing real demonstrations.} Can the generated videos match all-real policy training, and do they outperform dense-generated mixtures at the same mix ratio? \textbf{(iii) Ablation studies.} How does each component in \textsc{SKIP} affect performance?

\subsection{Experimental setup}
\label{sec:exp_setup}

\textbf{Benchmarks.}
We evaluate \textsc{SKIP} on the LIBERO simulation benchmark~\citep{liu2023libero} and on a four-task Franka Panda real-robot platform with teleoperated demonstrations. On LIBERO we train \textsc{SKIP-Generator} and \textsc{SKIP-Reconstructor} on the \textsc{SKIP-Selector} keyframes extracted from LIBERO-90, and evaluate on the four held-out suites.

\textbf{Baselines and conditions.}
We compare against three categories of baselines, namely Uniform, RDP~\citep{douglas1973algorithms}, and TriPSS~\citep{cakmak2025tripss} for keyframe selection, Wan~2.2-TI2V-5B~\citep{wan2025wan} for fully dense video generation, and action-free FILM~\citep{reda2022film} for interpolation. The five $\pi_{0.5}$~\citep{black2025pi_} policy training conditions share identical hyperparameters and differ only in data composition. Real uses all real demonstrations, SKIP-Mix70 and SKIP-Mix100 use $70\%$ and $100\%$ \textsc{SKIP}-generated videos with the remainder real, and Dense-Mix70 and Dense-Mix100 replace the synthetic portion with dense Wan~2.2 rollouts at the same ratios. We defer detailed preprocessing, splits, hyperparameters, hardware, and rollout protocols to Appendix~\ref{app:exp_protocol}.

\textbf{Evaluation Metrics.}
Since \textsc{SKIP-Selector} is unsupervised, supervised keyframe metrics such as F1 do not apply~\citep{otani2019rethinking}. We instead measure keyframe quality along two axes that are also robust to the uniform-sampling bias of linear-interpolation reconstruction, namely \emph{event coverage} via GripCov and MEC, and \emph{semantic span} via MaxSemDist and P95SemDist, summarized by a single operation-aware score $\mathrm{OAS}=\tfrac{1}{4}[\mathrm{GripCov}+\mathrm{MEC}+(1-\mathrm{MaxSemDist})+(1-\mathrm{P95SemDist})]\in[0,1]$.
Generated videos are evaluated using PSNR, SSIM~\citep{wang2004image}, LPIPS~\citep{zhang2018unreasonable}, and FVD~\citep{unterthiner2018towards}. Downstream utility is measured by $\pi_{0.5}$ closed-loop success rate. We provide exact metric definitions in Appendix~\ref{app:exp_protocol}, and verify that OAS tracks downstream success rate in Fig.~\ref{fig:oas_sr_corr} of Appendix~\ref{app:additional}.

\subsection{Sparse-to-dense generation}
\label{sec:exp_kf}

We evaluate the sparse-to-dense pipeline along three dimensions, namely the quality of the selected keyframes, the quality and cost of the reconstructed dense rollout, and the sensitivity to the keyframe budget $K_{\mathrm{kf}}$.

\textbf{Selection quality.}
We first evaluate whether \textsc{SKIP-Selector} captures the structural state changes that baselines often miss. Uniform and RDP test geometric selection without an event-aware mechanism, while TriPSS tests visual clustering without temporal order.

\begin{table}[b]
  \centering
  \caption{\textbf{Keyframe selection quality across the four LIBERO test suites at $K_{\mathrm{kf}}{=}41$}.}
  \label{tab:exp1_main}
  \scriptsize
  \begin{tabular*}{\linewidth}{@{\extracolsep{\fill}}lccccc}
    \toprule
    Method & MEC$\uparrow$ & GripCov$\uparrow$ & MaxSemDist$\downarrow$ & P95SemDist$\downarrow$ & \textbf{OAS}$\uparrow$ \\
    \midrule
    Uniform        & 0.779 & 0.771 & 0.137 & 0.111 & 0.825 \\
    RDP            & 0.855 & 0.864 & 0.189 & 0.137 & 0.848 \\
    TriPSS         & 0.702 & 0.711 & 0.145 & 0.117 & 0.787 \\
    \textbf{SKIP-Selector (Ours)} & \cellcolor{skipblue}\textbf{0.888} & \cellcolor{skipblue}\textbf{0.999} & \cellcolor{skipblue}\textbf{0.132} & \cellcolor{skipblue}\textbf{0.109} & \cellcolor{skipblue}\textbf{0.911} \\
    \bottomrule
  \end{tabular*}
\end{table}
As shown in Tab.~\ref{tab:exp1_main}, \textsc{SKIP-Selector} leads on every metric, with the largest margin on GripCov and a $+0.063$ OAS gap. We attribute this to the combination of multimodal fusion and explicit gripper-event protection, which jointly preserve the critical manipulation events. Per-event-count breakdowns (Fig.~\ref{fig:app_kf_bars}), event-hit visualizations (Fig.~\ref{fig:app_kf_viz}), and per-stream ablations (Tab.~\ref{tab:ablation_combined}(a)) are in Appendix~\ref{app:additional}.

\textbf{Generated video quality.}
We then evaluate the pixel-level and feature-level quality of the dense rollout. The keyframe-selector baselines share the same flow interpolator with \textsc{SKIP}, while Wan~2.2 replaces the entire pipeline with chunked dense generation.

\begin{table}[t]
  \centering
  \caption{\textbf{End-to-end video quality on LIBERO test suites (averaged across four suites):} Per-suite breakdown in Tab.~\ref{tab:app_bench_quality_persuite} of Appendix~\ref{app:diag_sparsetodense}.}
  \label{tab:bench_quality}
  \scriptsize
  \begin{tabular*}{\linewidth}{@{\extracolsep{\fill}}lcccc}
    \toprule
    Method & PSNR$\uparrow$ & SSIM$\uparrow$ & LPIPS$\downarrow$ & FVD$\downarrow$ \\
    \midrule
    Wan~2.2 (chunked) & 21.423 & 0.832 & 0.162 & 4177 \\
    Uniform & 21.345 & 0.847 & 0.130 & 716 \\
    RDP & 21.076 & 0.843 & 0.137 & 724 \\
    TriPSS & 21.033 & 0.827 & 0.157 & 640 \\
    \textbf{\textsc{SKIP} (Ours)} & \cellcolor{skipblue}\textbf{21.635} & \cellcolor{skipblue}\textbf{0.856} & \cellcolor{skipblue}\textbf{0.119} & \cellcolor{skipblue}\textbf{458} \\
    \bottomrule
  \end{tabular*}
\end{table}

As shown in Tab.~\ref{tab:bench_quality}, \textsc{SKIP} leads on all metrics, and on every suite-metric cell in the per-suite breakdown (Tab.~\ref{tab:app_bench_quality_persuite}). The dense Wan~2.2 baseline produces FVD $3\times$ to $15\times$ higher than any sparse-anchor pipeline, indicating that sparse-to-dense generation avoids the long-horizon drift accumulated by chunked recursive sampling. Qualitative success and failure rollouts are in Fig.~\ref{fig:app_gen_viz}.

\begin{table}[t]
  \centering
  \caption{\textbf{End-to-end inference time on LIBERO:} We report seconds per trajectory on one H20 GPU, where both methods use $50$ denoising steps and bf16 precision.}
  \label{tab:efficiency}
  \scriptsize
  \begin{tabular*}{\linewidth}{@{\extracolsep{\fill}}lcccc|cc}
    \toprule
    \multirow{2}{*}{Suite} & \multicolumn{4}{c|}{\textbf{\textsc{SKIP} (Ours)}} & \multirow{2}{*}{\makecell{Wan~2.2 \\(chunked)~$\downarrow$}} & \multirow{2}{*}{\textbf{Speedup}~$\uparrow$} \\
    \cmidrule(lr){2-5}
    & Keyframe gen.\ & Gap pred.\ & Interp.\ & \textbf{Total}~$\downarrow$ & & \\
    \midrule
    LIBERO-10      & 34.78 & 0.23 & 4.78 & \cellcolor{skipblue}\textbf{39.79} & 243.86 & \cellcolor{skipblue}\textbf{6.13$\times$} \\
    LIBERO-Goal    & 35.24 & 0.02 & 2.47 & \cellcolor{skipblue}\textbf{37.73} & 144.66 & \cellcolor{skipblue}\textbf{3.83$\times$} \\
    LIBERO-Object  & 35.45 & 0.02 & 2.28 & \cellcolor{skipblue}\textbf{37.75} & 137.92 & \cellcolor{skipblue}\textbf{3.65$\times$} \\
    LIBERO-Spatial & 34.68 & 0.02 & 1.94 & \cellcolor{skipblue}\textbf{36.64} & 124.74 & \cellcolor{skipblue}\textbf{3.40$\times$} \\
    \midrule
    \textbf{Weighted avg.}\!\! & 35.06 & 0.06 & 2.74 & \cellcolor{skipblue}\textbf{37.86} & 157.37 & \cellcolor{skipblue}\textbf{4.16$\times$} \\
    \bottomrule
  \end{tabular*}
\end{table}

\textbf{Efficiency.}
As shown in Tab.~\ref{tab:efficiency}, \textsc{SKIP} achieves a $4.16\times$ average speedup over recursive dense Wan~2.2 generation. The cost is dominated by keyframe generation at $\sim 35$~s per trajectory, while gap prediction and interpolation together add only $\sim 3$~s.

\textbf{Budget configurability.}
At the keyframe-extraction stage, $K_{\mathrm{kf}}$ can be set by the user or auto-selected per trajectory via the BIC criterion. Sweeping $K_{\mathrm{kf}}\in\{9,17,33,41,57\}$ in Appendix~\ref{app:additional} produces the cost-quality curve in Fig.~\ref{fig:app_density}, with an elbow at $K_{\mathrm{kf}}{=}41$ matching the median BIC-selected count on LIBERO-90. We therefore fix $K_{\mathrm{kf}}{=}41$ across the rest of the pipeline in this work, reflecting the data's intrinsic event density rather than manual tuning.

\subsection{Replacing real demonstrations}
\label{sec:exp_downstream}

Video metrics alone do not establish whether generated rollouts can serve as effective policy-training data. We therefore train $\pi_{0.5}$ policies with identical hyperparameters and vary only the data composition.

\begin{figure}[t]
  \centering
  \includegraphics[width=\linewidth]{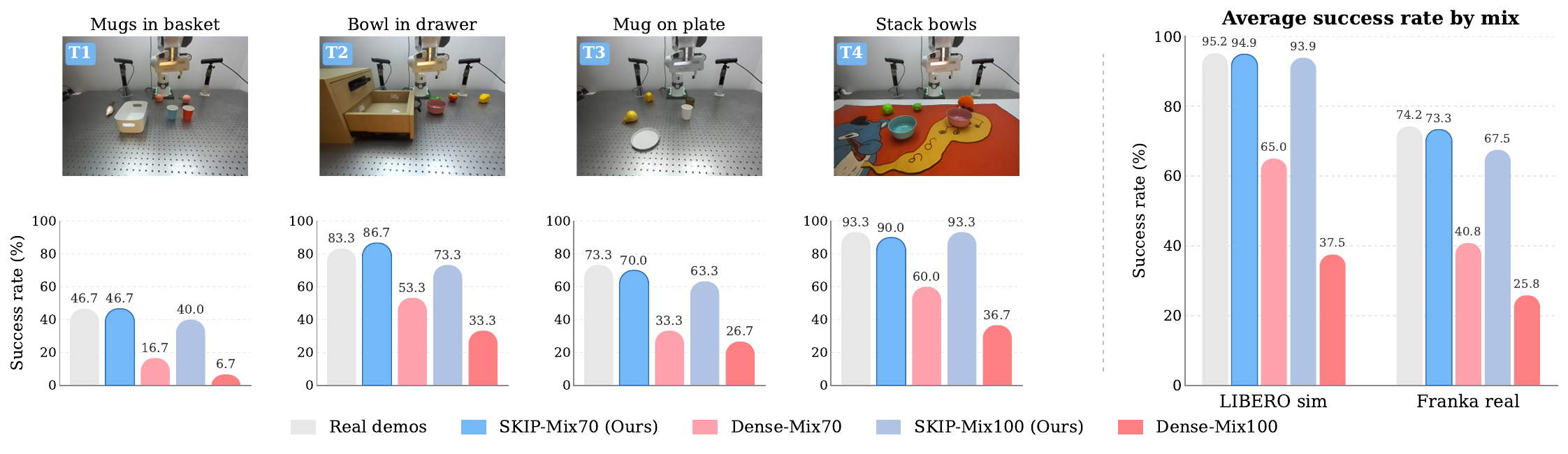}
  \caption{\textbf{$\pi_{0.5}$ success rate across five training mixes.} \textbf{Left:} screenshots of the four Franka Panda real-robot tasks T1 to T4 with their names labelled above, and per-task success rate over $30$ rollouts shown as bar charts below each task. \textbf{Right:} average success rate on LIBERO simulation and on the Franka real robot.}
  \label{fig:realrobot_sr}
\end{figure}

\textbf{Matching all-real training.}
As shown in Fig.~\ref{fig:realrobot_sr}, SKIP-Mix70 matches all-real training within $1$ pp on both platforms, costing $0.28$ pp on LIBERO and $0.84$ pp on the real robot. Even with all real data removed, SKIP-Mix100 drops only another $\sim 1$ pp in simulation, showing that \textsc{SKIP} videos alone can sustain near all-real policy performance.

\textbf{Drift collapse on dense baselines.}
Against the matched Dense-Mix baselines, SKIP-Mix70 outperforms Dense-Mix70 by $29.9$ pp in simulation and $32.5$ pp on the real robot, and the gap widens to $56.4$ and $41.7$ pp at $100\%$ synthetic. The drift accumulated by dense $49$-frame stitching propagates into the policy, so it is high-fidelity sparse generation, not the mere presence of synthetic video, that enables \textsc{SKIP} rollouts to substitute for real demonstrations. Per-suite and per-task breakdowns are in Tab.~\ref{tab:pi05_libero_sr} and Tab.~\ref{tab:realrobot_sr}.

\subsection{Ablation studies}
\label{sec:ablation}

We ablate two design groups in \textsc{SKIP}, namely the keyframe-selection design in \textsc{SKIP-Selector} and the action conditioning in \textsc{SKIP-Reconstructor}. Each ablation uses the same downstream $\pi_{0.5}$ success-rate protocol as \S\ref{sec:exp_downstream}, with results in Tab.~\ref{tab:ablation_downstream}.

\begin{table}[t]
  \centering
  \caption{\textbf{Downstream $\pi_{0.5}$ ablations of \textsc{SKIP} design components:} Gripper is the gripper-event post-processing in \textsc{SKIP-Selector} and Action is action conditioning in the interpolator.}
  \label{tab:ablation_downstream}
  \scriptsize
  \begin{tabular*}{\linewidth}{@{\extracolsep{\fill}}ccccccc}
    \toprule
    \multicolumn{4}{c}{Keyframe selection} & \multicolumn{1}{c}{Recovery} & \multicolumn{2}{c}{Avg. SR (\%)~$\uparrow$} \\
    \cmidrule(lr){1-4}\cmidrule(lr){5-5}\cmidrule(lr){6-7}
    Visual & Semantic & Proprioceptive & Gripper & Action & Sim & Real \\
    \midrule
    \cmark & \xmark & \xmark & \xmark & \cmark & 71.5 & 43.3 \\
    \xmark & \cmark & \xmark & \xmark & \cmark & 68.7 & 36.7 \\
    \xmark & \xmark & \cmark & \xmark & \cmark & 73.6 & 50.0 \\
    \cmark & \cmark & \xmark & \xmark & \cmark & 81.2 & 53.3 \\
    \cmark & \cmark & \cmark & \xmark & \cmark & 91.4 & 66.7 \\
    \cmark & \cmark & \cmark & \cmark & \xmark & 90.4 & 66.7 \\
    \rowcolor{skipblue}\cmark & \cmark & \cmark & \cmark & \cmark & \textbf{94.9} & \textbf{73.3} \\
    \bottomrule
  \end{tabular*}
\end{table}

\textbf{Keyframe selection.}
As shown in Tab.~\ref{tab:ablation_downstream}, no single stream suffices. Multimodal fusion progressively lifts performance, and gripper-event post-processing yields a further gain. The pattern suggests that smooth multimodal fusion captures the coarse manipulation structure, while explicit gripper-event protection safeguards the short discrete events that decide successful grasps. Per-stream selection-quality breakdowns are in Tab.~\ref{tab:ablation_combined}(a).

\textbf{Action conditioning.}
As shown in Tab.~\ref{tab:ablation_downstream}, AC-FILM improves both simulation and real-robot success over action-free FILM. The corresponding per-trajectory PSNR/SSIM/LPIPS improvements on dense recovery are reliable under a paired $t$-test at $p<0.01$, with full breakdowns in Tab.~\ref{tab:ablation_combined}(b). Action conditioning supplies the physical prior that matters around occlusions and contact transitions, where visual endpoints alone underconstrain robot motion. We further ablate the temporal segmentation choice against four alternatives (Uniform partition, Spectral clustering~\citep{ng2001spectral}, Agglomerative clustering~\citep{ward1963hierarchical}, Bayesian change-point~\citep{adams2007bayesian}) in \S\ref{app:diag_sparsetodense}, and the fusion weighting against three strategies (effective-rank-weighted, fixed-weight, grid search) in \S\ref{app:diag_ablation}. Both confirm KTS and equal weighting as the strongest configurations on every metric reported.

%===============================================================================
\section{Conclusion}
\label{sec:conclusion}

We propose \textsc{SKIP}, a sparse keyframe interpolation framework for embodied world models that selects event-preserving keyframes through robot-aware multimodal fusion, generates only those keyframes with a video diffusion model, and reconstructs dense action-aligned rollouts via a learned gap predictor and an action-conditioned interpolator. On the LIBERO benchmark and a four-task Franka Panda real-robot platform, \textsc{SKIP} produces dense rollouts $4.16\times$ faster than a dense Wan~2.2 baseline while reducing aggregate FVD by $89.0\%$, and \textsc{SKIP} videos are effective policy-training data. Even fully replacing real demonstrations drops $\pi_{0.5}$ success by only $1.3$ pp in simulation and $6.7$ pp on the real robot, whereas fully dense-generated rollouts collapse by $48$ to $58$ pp. These results suggest that, on the manipulation benchmarks studied, much of a dense rollout can be recovered from a few decisive manipulation events, opening a practical path toward scalable visual data for robot policy learning and applications such as planning preview and behavior-level inspection.

%===============================================================================
\section{Limitations}
\label{sec:limitations}

Our method assumes a fixed third-person camera and a single-arm manipulator, and has been evaluated on a limited set of datasets. Its clearest failure mode is on long compound episodes, where errors near phase switches accumulate in the dense reconstruction, which explicit phase-aware recovery could address.
We plan to extend our approach in three directions. First, we will evaluate on more datasets and policy backbones to test general effectiveness. Second, we will adapt the event-driven keyframe strategy to multi-view and bimanual manipulation. Third, we will train vision-language-action models on keyframes alone rather than dense video, testing whether sparse event-aware training preserves policy performance at lower compute.

%===============================================================================
\clearpage
\bibliography{references}

%===============================================================================
\clearpage
\appendix

\noindent\textbf{Appendix overview.} Appendix~\ref{app:exp_protocol} details the datasets, splits, baselines, and evaluation metrics. Appendix~\ref{app:impl} reports hardware, training hyperparameters, and feature-stream specifications. Appendix~\ref{app:additional} provides supplementary diagnostics extending the main experiments, organized into three subsections that follow the main paper's experimental flow, namely sparse-to-dense generation (\S\ref{sec:exp_kf}), policy-data results (\S\ref{sec:exp_downstream}), and ablation studies (\S\ref{sec:ablation}).

\section{Datasets and evaluation protocol}
\label{app:exp_protocol}

Tab.~\ref{tab:app_datasets} lists the source datasets and trajectory counts, and Fig.~\ref{fig:realrobot_tasks} visualizes the four real-robot tasks.

\begin{table}[!htbp]
  \centering
  \caption{\textbf{Datasets at a glance.} All LIBERO numbers are no-op-filtered trajectory counts. The four real-robot tasks total 192 demonstrations.}
  \label{tab:app_datasets}
  \scriptsize
  \begin{tabular*}{\linewidth}{@{\extracolsep{\fill}}llr}
    \toprule
    Dataset & Source & \# Trajectories \\
    \midrule
    libero\_90      & LIBERO              & 3528 \\
    \midrule
    libero\_10      & LIBERO              & 301 \\
    libero\_goal    & LIBERO              & 376 \\
    libero\_object  & LIBERO              & 432 \\
    libero\_spatial & LIBERO              & 394 \\
    \midrule
    T1: put both the red mug and the blue mug in the basket  & Franka Panda teleop & 48 \\
    T2: put the red bowl in the drawer and close the drawer  & Franka Panda teleop & 48 \\
    T3: put the white mug on the plate                       & Franka Panda teleop & 48 \\
    T4: stack the red bowl on the green bowl                 & Franka Panda teleop & 48 \\
    \bottomrule
  \end{tabular*}
\end{table}

\subsection{LIBERO simulation}

LIBERO contains five suites totalling 5031 trajectories after no-op filtering.
Raw episodes are unified into a trajectory format with RGB frames, actions, proprioception, and language instructions following~\citep{li2025keyworld}, and the no-op-filtered version (semantically equivalent to the official \texttt{*\_no\_noops} RLDS preprocessing) is used after discarding dummy episodes.
RGB frames are bicubic-resized from $512{\times}512$ to $256{\times}256$, and $\pi_{0.5}$ further pads them to $224{\times}224$ for the PaliGemma encoder.
The raw 9-D LIBERO state (gripper $\times 2$, end-effector position, quaternion) is converted to an 8-D vector via axis-angle representation, and actions remain 7-D continuous commands.

\textbf{Splits.}
libero\_90 (3528 traj) is the training pool for the full SKIP pipeline, namely SKIP-Selector keyframe extraction, SKIP-Generator fine-tuning, and SKIP-Reconstructor training all run on libero\_90.
The other four suites (libero\_10 / goal / object / spatial, 1503 traj) are held out for SKIP video-quality evaluation, reported in Tab.~\ref{tab:exp1_main} through Tab.~\ref{tab:efficiency}.
The same four suites also serve as the source for downstream $\pi_{0.5}$ studies, providing the $1503$ trajectories on which the five training conditions defined in \S\ref{sec:exp_setup} (Real demos, SKIP-Mix70/100, Dense-Mix70/100) are built. Each policy is evaluated via closed-loop rollouts on the same four suites.

\subsection{Real-robot platform}

\begin{figure}[!htbp]
  \centering
  \includegraphics[width=\linewidth]{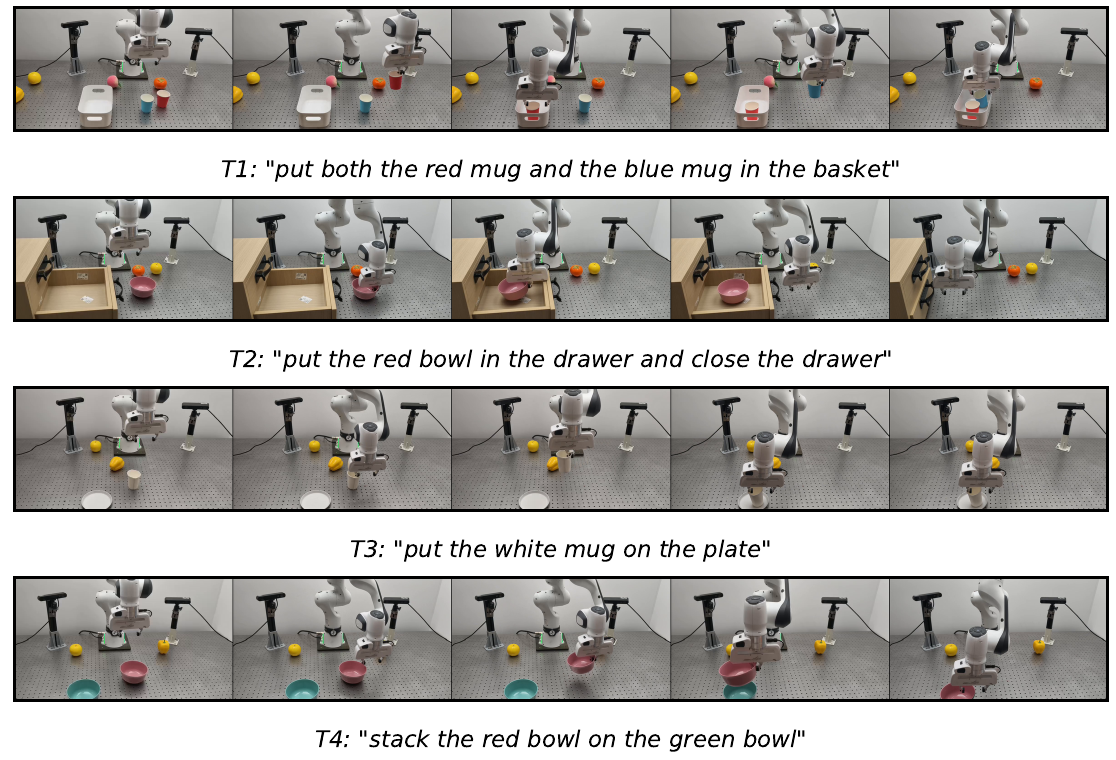}
  \caption{\textbf{Real-robot task overview.} Each row corresponds to one task and matches the T1--T4 columns of Tab.~\ref{tab:realrobot_sr}.}
  \label{fig:realrobot_tasks}
\end{figure}

The platform is a $7$-DoF Franka Emika Panda with a single third-person RGB camera.
Fig.~\ref{fig:realrobot_tasks} shows the four manipulation tasks, namely sequential pick-and-place (T1), pick-and-place plus drawer operation (T2), single pick-and-place (T3), and object stacking (T4). We collect 48 teleoperated demonstrations per task, totalling 192 episodes.
$\pi_{0.5}$ training uses the same five conditions as in simulation (Real demos, SKIP-Mix70/100, Dense-Mix70/100), with the SKIP synthetic portion produced by a SKIP pipeline separately fine-tuned on real-robot demonstrations.
Evaluation is 30 closed-loop rollouts per task with initial placements randomized within a fixed range.

\subsection{Baselines and metrics}

\textbf{Keyframe baselines} all use the same budget $K_{\mathrm{kf}}{=}41$. Uniform selects equispaced frames, RDP~\citep{douglas1973algorithms} follows the KeyWorld~\citep{li2025keyworld} choice, and TriPSS~\citep{cakmak2025tripss} applies HDBSCAN over three visual streams.
The \textbf{dense video baseline} is Wan~2.2-TI2V-5B~\citep{wan2025wan} fine-tuned on dense $49$-frame chunks of LIBERO-90 with the same recipe and training pool as \textsc{SKIP-Generator}, run recursively chunk by chunk at synthesis time. This matches \textsc{SKIP}'s training data and protocol, isolating the effect of sparse versus dense generation from any data-exposure or training-recipe asymmetry.
The \textbf{interpolation baseline} is action-free FILM~\citep{reda2022film} fine-tuned on the same SKIP-Selector keyframes with the same recipe as AC-FILM.

\textbf{Keyframe-quality metrics.}
We avoid supervised metrics such as F1, which are known to be sensitive to video pre-segmentation rather than to the keyframe algorithm itself~\citep{otani2019rethinking}, and instead use unsupervised coverage measures.
GripCov is the fraction of gripper sign-change events covered within $\pm 2$ frames.
MEC applies the same $\pm 2$-frame coverage to four event types (gripper sign change, joint-velocity spike, trajectory turning point, acceleration burst).
MaxSemDist is the maximum cosine distance between adjacent keyframes' DINOv3 features and P95SemDist its 95th percentile.
\textbf{Video quality} uses PSNR, SSIM, and LPIPS at $256{\times}256$, plus FVD via an I3D backbone.
\textbf{Downstream utility} is measured by $\pi_{0.5}$ closed-loop success rate.

\section{Implementation Details}
\label{app:impl}

\subsection{Training and inference configuration}
\label{app:impl_train}

Table~\ref{tab:app_impl} lists every training and inference configuration for reproducibility, including the System, \textsc{SKIP-Generator}, Gap predictor, AC-FILM, and $\pi_{0.5}$ fine-tune blocks.

\begin{table}[!htbp]
  \centering
  \caption{\textbf{Implementation details.}}
  \label{tab:app_impl}
  \scriptsize
  \begin{tabular*}{\linewidth}{@{\extracolsep{\fill}}lll}
    \toprule
    Module & Element & Detail \\
    \midrule
    \multirow{5}{*}{System}
        & OS                 & Ubuntu 22.04 \\
        & CUDA               & 12.1 \\
        & Python             & 3.10 \\
        & PyTorch            & 2.4.1 \\
        & Device             & $8\times$ NVIDIA H20 \\
    \midrule
    \multirow{7}{*}{\textsc{SKIP-Generator}}
        & Batch size         & 1 \\
        & Epochs             & 30 \\
        & Resolution         & $256{\times}256$ \\
        & Optimizer          & AdamW \\
        & Learning rate      & 1e-5 \\
        & Precision          & bf16 \\
        & Trainable modules  & DiT only (VAE/text-encoder frozen) \\
    \midrule
    \multirow{5}{*}{Gap predictor}
        & Encoder            & frozen ResNet-50~\citep{he2016deep} \\
        & Batch size         & 8 \\
        & Epochs             & 100 \\
        & Optimizer          & Adam \\
        & Learning rate      & 3e-4 \\
    \midrule
    \multirow{7}{*}{AC-FILM}
        & Batch size         & 4 \\
        & Epochs             & 20 (1 backbone-frozen + 19 joint) \\
        & Optimizer          & AdamW \\
        & Learning rate      & 2e-4 \\
        & Weight decay       & 1e-5 \\
        & Loss               & $\ell_1$ + VGG perceptual + Gram style \\
        & FiLM proj.\ init   & Zero (identity at start) \\
    \midrule
    \multirow{5}{*}{$\pi_{0.5}$ fine-tune}
        & Batch size         & 32 \\
        & Steps              & 30\,000 \\
        & Optimizer          & AdamW (cosine schedule) \\
        & Learning rate      & 5e-5 (peak) \\
        & Precision          & bf16 + gradient checkpointing \\
    \bottomrule
  \end{tabular*}
\end{table}

\subsection{Multimodal feature streams}
\label{app:impl_streams}

The three feature streams that feed \textsc{SKIP-Selector} are described below. All streams are computed offline on the full dataset and cached before \textsc{SKIP-Selector} runs.

\textbf{Visual stream (V).}
We compute UniMatch~\citep{xu2023unifying} optical flow between consecutive frames, then summarize each flow field with a histogram-of-flow (HoF) descriptor following~\citet{wang2013dense}: a $2{\times}2$ spatial grid, $8$ orientation bins, and a low-magnitude (`zero-flow') bin, yielding $4{\times}9=36$ dimensions.
The last frame's descriptor is duplicated to align with the image count.

\textbf{Semantic stream (S).}
DINOv3 ViT-L/16~\citep{simeoni2025dinov3} features are extracted at $224{\times}224$ with ImageNet normalization.
We concatenate the CLS token and the mean of the patch tokens, producing a 2048-D vector, and L2-normalize.

\textbf{Proprioceptive stream (P).}
Six feature groups, 41 dimensions in total: raw state (8), first-order velocity (8), second-order acceleration (8), multi-scale RDP perpendicular distances at window radii $\{3,5,9,15,25\}$ in both the position subspace and the full state ($5{\times}2=10$), multi-scale Menger curvatures of the position trajectory (5), and gripper change rate plus binary open/close event (2).
All states are first converted to axis-angle for rotation, with a sequential unwrapping step to handle the $\theta_r=\pi$ discontinuity of quaternion-to-axis-angle conversion.
Each group is z-score normalized before concatenation.

\section{Supplementary diagnostics}
\label{app:failure_decomp}
\label{app:additional}

This section extends the main experiments of \S\ref{sec:exp} in order. We first add analyses of the sparse-to-dense generation experiments from \S\ref{sec:exp_kf}, then breakdowns of the policy-data experiments from \S\ref{sec:exp_downstream}, and finally intermediate-metric ablations for \S\ref{sec:ablation}.

\subsection{Sparse-to-dense generation extensions (\S\ref{sec:exp_kf})}
\label{app:diag_sparsetodense}

\paragraph{Selection-quality details.}

\begin{figure}[!htbp]
  \centering
  \includegraphics[width=\linewidth]{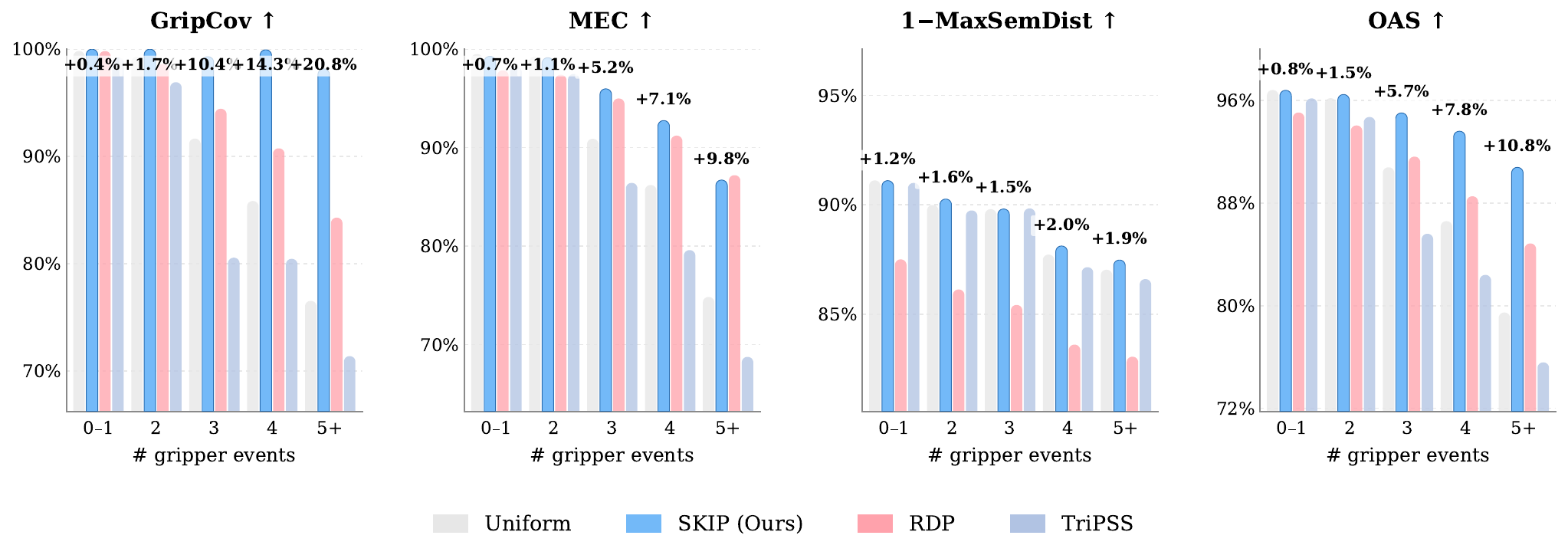}
  \caption{\textbf{Per-event-count keyframe-quality breakdown.} Held-out LIBERO trajectories are bucketed by gripper-event count (0--1, 2, 3, 4, 5+), with each panel reporting a different keyframe-quality metric. \textsc{SKIP-Selector}'s margin over Uniform, RDP, and TriPSS widens as manipulation complexity grows, with the largest gains on the $5{+}$-event bucket.}
  \label{fig:app_kf_bars}
\end{figure}

\textbf{Per-event-count keyframe-quality breakdown.}
As shown in Fig.~\ref{fig:app_kf_bars}, binning held-out LIBERO trajectories by gripper-event count per episode reveals that \textsc{SKIP-Selector}'s GripCov stays above $0.99$ on every bin including the $5{+}$-event bin, while Uniform's GripCov drops sharply. The margin over Uniform, RDP, and TriPSS widens monotonically with manipulation complexity.

\begin{figure}[!htbp]
  \centering
  \includegraphics[width=\linewidth]{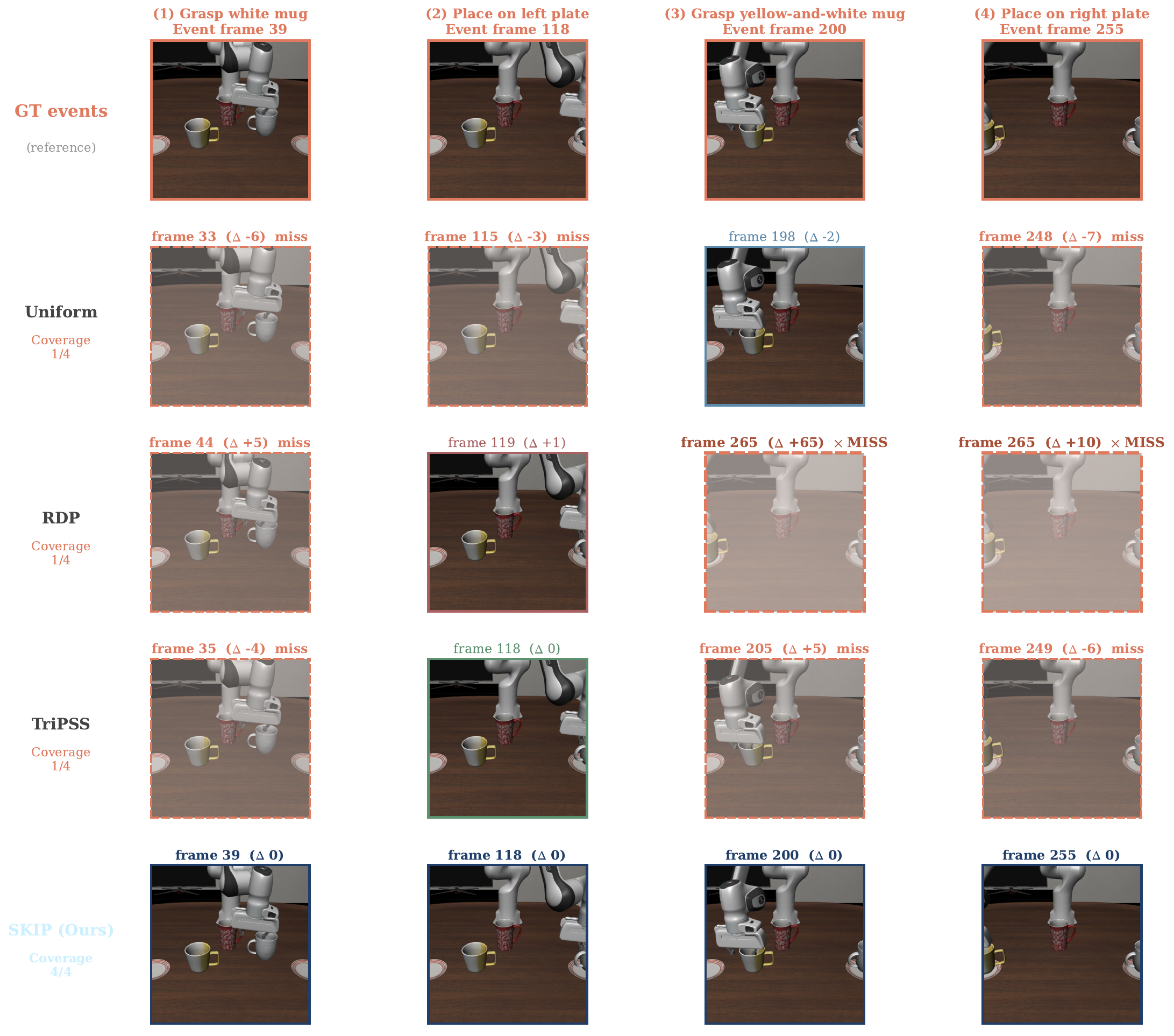}
  \caption{\textbf{Per-trajectory event-hit visualization of keyframe selection.} A LIBERO trajectory with four ground-truth gripper events. For each method, the column shows the chosen keyframe closest to each event with its frame offset and any missed event. \textsc{SKIP-Selector} (bottom row) hits all four events exactly, while Uniform, RDP, and TriPSS each miss multiple events.}
  \label{fig:app_kf_viz}
\end{figure}

\textbf{Per-trajectory event-hit visualization.}
As shown in Fig.~\ref{fig:app_kf_viz}, on a representative LIBERO trajectory with four ground-truth gripper events, \textsc{SKIP-Selector} hits all four exactly while Uniform, RDP, and TriPSS each miss multiple events.

\textbf{Per-suite selection-quality breakdown.}
While main-body Tab.~\ref{tab:exp1_main} reports only across-suite averages, the per-suite breakdown in Tab.~\ref{tab:app_sel_per_suite} exposes whether \textsc{SKIP-Selector}'s average advantage is uniform across LIBERO suites or driven by specific ones. \textsc{SKIP-Selector} strictly outperforms every baseline on every (suite, metric) cell, because the multimodal fusion combines event-coverage signals (MEC, GripCov) with semantic-span signals (MaxSemDist, P95SemDist) that no single-modality baseline can match across the diverse task distributions of LIBERO. The AVG block at the bottom matches Tab.~\ref{tab:exp1_main}.

\begin{table}[!htbp]
  \centering
  \caption{\textbf{Per-suite selection-quality breakdown.} AVG block at bottom matches Tab.~\ref{tab:exp1_main}.}
  \label{tab:app_sel_per_suite}
  \scriptsize
  \begin{tabular*}{\linewidth}{@{\extracolsep{\fill}}llccccc}
    \toprule
    Suite & Method & MEC$\uparrow$ & GripCov$\uparrow$ & MaxSemDist$\downarrow$ & P95SemDist$\downarrow$ & \textbf{OAS}$\uparrow$ \\
    \midrule
    \multirow{4}{*}{LIBERO-10} & Uniform & 0.755 & 0.748 & 0.157 & 0.125 & 0.805 \\
    & RDP & 0.838 & 0.846 & 0.213 & 0.150 & 0.830 \\
    & TriPSS & 0.685 & 0.694 & 0.160 & 0.128 & 0.773 \\
    & \textbf{SKIP-Selector (Ours)} & \cellcolor{skipblue}\textbf{0.872} & \cellcolor{skipblue}\textbf{0.997} & \cellcolor{skipblue}\textbf{0.152} & \cellcolor{skipblue}\textbf{0.123} & \cellcolor{skipblue}\textbf{0.899} \\
    \midrule
    \multirow{4}{*}{LIBERO-Goal} & Uniform & 0.788 & 0.781 & 0.135 & 0.110 & 0.831 \\
    & RDP & 0.860 & 0.870 & 0.184 & 0.135 & 0.853 \\
    & TriPSS & 0.706 & 0.714 & 0.142 & 0.114 & 0.791 \\
    & \textbf{SKIP-Selector (Ours)} & \cellcolor{skipblue}\textbf{0.890} & \cellcolor{skipblue}\textbf{0.999} & \cellcolor{skipblue}\textbf{0.128} & \cellcolor{skipblue}\textbf{0.108} & \cellcolor{skipblue}\textbf{0.913} \\
    \midrule
    \multirow{4}{*}{LIBERO-Object} & Uniform & 0.770 & 0.760 & 0.147 & 0.115 & 0.817 \\
    & RDP & 0.847 & 0.857 & 0.198 & 0.142 & 0.841 \\
    & TriPSS & 0.694 & 0.702 & 0.155 & 0.118 & 0.781 \\
    & \textbf{SKIP-Selector (Ours)} & \cellcolor{skipblue}\textbf{0.880} & \cellcolor{skipblue}\textbf{0.999} & \cellcolor{skipblue}\textbf{0.141} & \cellcolor{skipblue}\textbf{0.113} & \cellcolor{skipblue}\textbf{0.906} \\
    \midrule
    \multirow{4}{*}{LIBERO-Spatial} & Uniform & 0.799 & 0.791 & 0.113 & 0.098 & 0.845 \\
    & RDP & 0.871 & 0.879 & 0.165 & 0.125 & 0.865 \\
    & TriPSS & 0.719 & 0.730 & 0.127 & 0.112 & 0.802 \\
    & \textbf{SKIP-Selector (Ours)} & \cellcolor{skipblue}\textbf{0.906} & \cellcolor{skipblue}\textbf{0.999} & \cellcolor{skipblue}\textbf{0.111} & \cellcolor{skipblue}\textbf{0.096} & \cellcolor{skipblue}\textbf{0.924} \\
    \midrule
    \multirow{4}{*}{\textbf{AVG}} & Uniform & 0.779 & 0.771 & 0.137 & 0.111 & 0.825 \\
    & RDP & 0.855 & 0.864 & 0.189 & 0.137 & 0.848 \\
    & TriPSS & 0.702 & 0.711 & 0.145 & 0.117 & 0.787 \\
    & \textbf{SKIP-Selector (Ours)} & \cellcolor{skipblue}\textbf{0.888} & \cellcolor{skipblue}\textbf{0.999} & \cellcolor{skipblue}\textbf{0.132} & \cellcolor{skipblue}\textbf{0.109} & \cellcolor{skipblue}\textbf{0.911} \\
    \bottomrule
  \end{tabular*}
\end{table}

\textbf{Segmenter selection quality.}
To isolate the contribution of KTS dynamic programming, we replace the temporal segmentation step with four alternatives while holding V/S/P fusion and gripper-event post-processing fixed. The alternatives span uniform partitioning (which ignores the affinity matrix), spectral clustering~\citep{ng2001spectral} ($k$-cut without temporal-order constraints), agglomerative clustering~\citep{ward1963hierarchical} (local similarity), and Bayesian online change-point detection~\citep{adams2007bayesian} (explicit temporal segments). As shown in Tab.~\ref{tab:app_segmenter_sel}, KTS leads on every selection-quality metric. The largest GripCov margin (KTS $0.999$ vs.\ Bayesian change-point $0.972$, a $0.027$ gap) reflects the dynamic-programming objective's ability to land boundaries precisely on event transitions rather than near them. Bayesian online change-point is the closest competitor on OAS ($0.891$ vs.\ KTS $0.911$) because it also leverages temporal order, but its locally-greedy segmentation does not optimize the global within-segment variance KTS minimizes. The three no-temporal-order baselines (Uniform / Spectral / Agglomerative) trail by between $0.033$ and $0.096$ OAS.

\begin{table}[!htbp]
  \centering
  \caption{\textbf{Alternative temporal segmenters: selection quality.}}
  \label{tab:app_segmenter_sel}
  \scriptsize
  \begin{tabular*}{\linewidth}{@{\extracolsep{\fill}}lccccc}
    \toprule
    Segmenter & MEC$\uparrow$ & GripCov$\uparrow$ & MaxSemDist$\downarrow$ & P95SemDist$\downarrow$ & \textbf{OAS}$\uparrow$ \\
    \midrule
    Uniform partition on $\mathbf{G}$ & 0.748 & 0.815 & 0.162 & 0.142 & 0.815 \\
    Spectral clustering ($k$-cut) & 0.838 & 0.875 & 0.155 & 0.130 & 0.857 \\
    Agglomerative clustering & 0.832 & 0.952 & 0.149 & 0.124 & 0.878 \\
    Bayesian online change-point & 0.846 & 0.972 & 0.139 & 0.116 & 0.891 \\
    \rowcolor{skipblue} \textbf{KTS (Ours)} & \textbf{0.888} & \textbf{0.999} & \textbf{0.132} & \textbf{0.109} & \textbf{0.911} \\
    \bottomrule
  \end{tabular*}
\end{table}

\textbf{Segmenter downstream SR.}
As shown in Tab.~\ref{tab:app_segmenter_down}, the SR ranking follows the selection-quality ranking of Tab.~\ref{tab:app_segmenter_sel}, namely KTS $>$ Bayesian change-point $>$ Agglomerative $>$ Spectral $>$ Uniform on both LIBERO sim avg and Franka real avg. KTS-Mix100 reaches $93.9\%$ sim and $67.5\%$ real, beating the second-best Bayesian change-point by $5.9$~pp sim and $7.5$~pp real. The two no-temporal-order baselines (Uniform partition on $\mathbf{G}$ and Spectral clustering) trail by between $11.5$ and $17.7$~pp on sim and between $11.7$ and $18.3$~pp on real, confirming that explicit temporal segmentation is essential for downstream policy performance.

\begin{table}[!htbp]
  \centering
  \caption{\textbf{Alternative temporal segmenters: downstream $\pi_{0.5}$ SR (Mix100).} Each alternative pipes its keyframes through identical SKIP-Generator + SKIP-Reconstructor + $\pi_{0.5}$ training.}
  \label{tab:app_segmenter_down}
  \scriptsize
  \begin{tabular*}{\linewidth}{@{\extracolsep{\fill}}lccccccccccc}
    \toprule
    \multirow{2}{*}{Segmenter} & \multicolumn{5}{c}{LIBERO Sim SR (\%)} & \multicolumn{5}{c}{Franka Real (X/30)} \\
    \cmidrule(lr){2-6}\cmidrule(lr){7-11}
    & L-10 & L-Goal & L-Object & L-Spatial & \textbf{Avg} & T1 & T2 & T3 & T4 & \textbf{Avg (\%)} \\
    \midrule
    Uniform partition on $\mathbf{G}$ & 65.0 & 80.0 & 72.0 & 88.0 & 76.2 & 9/30 & 15/30 & 15/30 & 20/30 & 49.2 \\
    Spectral clustering & 70.0 & 85.0 & 80.0 & 94.6 & 82.4 & 10/30 & 18/30 & 17/30 & 22/30 & 55.8 \\
    Agglomerative clustering & 74.0 & 86.0 & 86.0 & 93.0 & 84.8 & 11/30 & 19/30 & 18/30 & 21/30 & 57.5 \\
    Bayesian change-point & 80.0 & 89.0 & 88.0 & 95.0 & 88.0 & 11/30 & 20/30 & 18/30 & 23/30 & 60.0 \\
    \rowcolor{skipblue} \textbf{KTS (Ours)} & \textbf{88.8} & \textbf{96.0} & \textbf{94.2} & \textbf{96.4} & \textbf{93.9} & \textbf{12/30} & \textbf{22/30} & \textbf{19/30} & \textbf{28/30} & \textbf{67.5} \\
    \bottomrule
  \end{tabular*}
\end{table}

\paragraph{Video-quality details.}

\textbf{Per-suite video quality.}
Tab.~\ref{tab:app_bench_quality_persuite} expands the suite-averaged Tab.~\ref{tab:bench_quality} of the main body into the four LIBERO suites. \textsc{SKIP} strictly wins every (suite, metric) cell. The dense Wan~2.2 baseline's per-suite FVD spans $2768$ to $6610$, $3\times$ to $15\times$ higher than any sparse-anchor row on each suite, confirming that chunked recursive sampling accumulates substantial long-horizon drift that sparse keyframe generation avoids.

\begin{table}[!htbp]
  \centering
  \caption{\textbf{Per-suite end-to-end video quality on LIBERO test suites.} Detailed breakdown of the averaged Tab.~\ref{tab:bench_quality} of the main body. \textsc{SKIP} leads on every (suite, metric) cell.}
  \label{tab:app_bench_quality_persuite}
  \scriptsize
  \begin{tabular*}{\linewidth}{@{\extracolsep{\fill}}llcccc}
    \toprule
    Suite & Method & PSNR$\uparrow$ & SSIM$\uparrow$ & LPIPS$\downarrow$ & FVD$\downarrow$ \\
    \midrule
    \multirow{5}{*}{LIBERO-10}
      & Wan~2.2 (chunked) & 21.373 & 0.833 & 0.176 & 3703 \\
      & Uniform           & 21.112 & 0.837 & 0.146 & 596 \\
      & RDP               & 21.004 & 0.836 & 0.149 & 562 \\
      & TriPSS            & 20.836 & 0.811 & 0.161 & 636 \\
      & \textbf{\textsc{SKIP} (Ours)} & \cellcolor{skipblue}\textbf{21.504} & \cellcolor{skipblue}\textbf{0.844} & \cellcolor{skipblue}\textbf{0.138} & \cellcolor{skipblue}\textbf{548} \\
    \midrule
    \multirow{5}{*}{LIBERO-Goal}
      & Wan~2.2 (chunked) & 20.819 & 0.826 & 0.160 & 3238 \\
      & Uniform           & 20.463 & 0.826 & 0.151 & 617 \\
      & RDP               & 20.479 & 0.825 & 0.153 & 572 \\
      & TriPSS            & 20.515 & 0.823 & 0.146 & 508 \\
      & \textbf{\textsc{SKIP} (Ours)} & \cellcolor{skipblue}\textbf{20.958} & \cellcolor{skipblue}\textbf{0.842} & \cellcolor{skipblue}\textbf{0.130} & \cellcolor{skipblue}\textbf{468} \\
    \midrule
    \multirow{5}{*}{LIBERO-Object}
      & Wan~2.2 (chunked) & 21.890 & 0.838 & 0.157 & 6610 \\
      & Uniform           & 22.085 & 0.877 & 0.098 & 845 \\
      & RDP               & 21.934 & 0.873 & 0.105 & 943 \\
      & TriPSS            & 21.787 & 0.831 & 0.147 & 774 \\
      & \textbf{\textsc{SKIP} (Ours)} & \cellcolor{skipblue}\textbf{22.371} & \cellcolor{skipblue}\textbf{0.885} & \cellcolor{skipblue}\textbf{0.090} & \cellcolor{skipblue}\textbf{445} \\
    \midrule
    \multirow{5}{*}{LIBERO-Spatial}
      & Wan~2.2 (chunked) & 21.525 & 0.833 & 0.157 & 2768 \\
      & Uniform           & 21.552 & 0.841 & 0.134 & 762 \\
      & RDP               & 20.759 & 0.831 & 0.146 & 751 \\
      & TriPSS            & 20.851 & 0.839 & 0.176 & 622 \\
      & \textbf{\textsc{SKIP} (Ours)} & \cellcolor{skipblue}\textbf{21.576} & \cellcolor{skipblue}\textbf{0.845} & \cellcolor{skipblue}\textbf{0.126} & \cellcolor{skipblue}\textbf{393} \\
    \bottomrule
  \end{tabular*}
\end{table}

\paragraph{Budget sensitivity.}

The keyframe budget $K_{\mathrm{kf}}$ controls the cost-quality trade-off of the sparse-to-dense paradigm. Lower $K_{\mathrm{kf}}$ saves generator inference cost but risks missing manipulation events, while higher $K_{\mathrm{kf}}$ approaches dense generation in cost without proportional quality gain. We sweep $K_{\mathrm{kf}}\in\{9,17,33,41,57\}$ and report selection metrics, video quality, and downstream policy SR at each point.

\begin{figure}[!htbp]
  \centering
  \includegraphics[width=\linewidth]{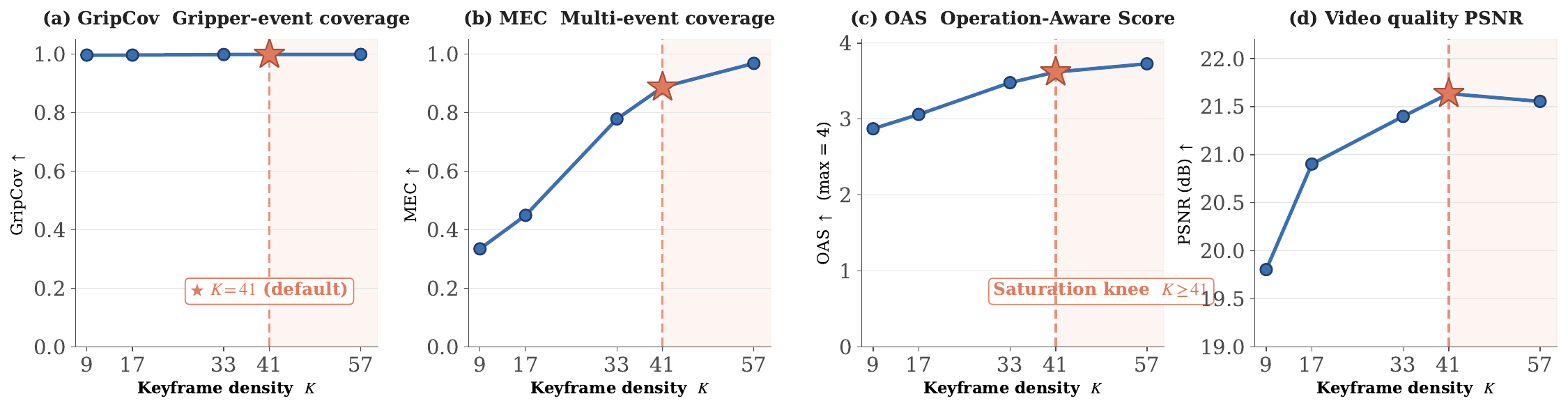}
  \caption{\textbf{Event coverage and video quality saturate near $K_{\mathrm{kf}}=41$.} Increasing the keyframe budget improves event coverage and downstream video quality up to the elbow near $K_{\mathrm{kf}}=41$. Higher budgets give smaller marginal gains while increasing inference cost.}
  \label{fig:app_density}
\end{figure}

\textbf{Saturation curve.}
As shown in Fig.~\ref{fig:app_density}, GripCov, MEC, and OAS all enter an elbow near $K_{\mathrm{kf}}=41$ on LIBERO, and downstream-generator PSNR enters its elbow at the same point. The choice of $K_{\mathrm{kf}}=41$ thus rests on three independent criteria, namely event-coverage saturation, downstream PSNR elbow, and BIC consensus on segment count, and aligns with both the $4N{+}1$ and $8N{+}1$ frame budgets of common video diffusion backbones.

\textbf{Selection and video metrics at each $K_{\mathrm{kf}}$.}
As shown in Tab.~\ref{tab:app_kkf_sel_vid}, all five selection-quality metrics improve monotonically with $K_{\mathrm{kf}}$ up to $K_{\mathrm{kf}}=41$ and then plateau or slightly regress at $K_{\mathrm{kf}}=57$ because the selector starts emitting near-redundant keyframes. PSNR / SSIM / LPIPS follow the same elbow profile, indicating that the video generator gains less per extra keyframe past $K_{\mathrm{kf}}=41$. The $K_{\mathrm{kf}}=33$ row remains within $0.013$ OAS and $0.24$~dB PSNR of $K_{\mathrm{kf}}=41$, suggesting that a $20\%$ keyframe-count reduction is available with only marginal quality loss.

\begin{table}[!htbp]
  \centering
  \caption{\textbf{Keyframe-budget sweep: selection + video metrics.}}
  \label{tab:app_kkf_sel_vid}
  \scriptsize
  \begin{tabular*}{\linewidth}{@{\extracolsep{\fill}}ccccccccc}
    \toprule
    $K_{\mathrm{kf}}$ & MEC$\uparrow$ & GripCov$\uparrow$ & MaxSemDist$\downarrow$ & P95SemDist$\downarrow$ & OAS$\uparrow$ & PSNR$\uparrow$ & SSIM$\uparrow$ & LPIPS$\downarrow$ \\
    \midrule
    9 & 0.742 & 0.852 & 0.205 & 0.176 & 0.803 & 19.804 & 0.812 & 0.158 \\
    17 & 0.818 & 0.934 & 0.165 & 0.142 & 0.861 & 20.902 & 0.832 & 0.139 \\
    33 & 0.870 & 0.978 & 0.140 & 0.116 & 0.898 & 21.398 & 0.847 & 0.126 \\
    \rowcolor{skipblue} \textbf{41 (Ours)} & \textbf{0.888} & \textbf{0.999} & \textbf{0.132} & \textbf{0.109} & \textbf{0.911} & \textbf{21.635} & \textbf{0.856} & \textbf{0.119} \\
    57 & 0.882 & 0.998 & 0.137 & 0.113 & 0.908 & 21.554 & 0.852 & 0.123 \\
    \bottomrule
  \end{tabular*}
\end{table}

\textbf{Downstream policy SR at each $K_{\mathrm{kf}}$.}
As shown in Tab.~\ref{tab:app_kkf_down}, downstream $\pi_{0.5}$ SR scales sub-linearly with $K_{\mathrm{kf}}$ and peaks at $K_{\mathrm{kf}}=41$ on both LIBERO sim (avg $93.9\%$) and Franka real (avg $67.5\%$). $K_{\mathrm{kf}}=57$ regresses by $0.6$~pp sim and $4.2$~pp real, with FVD up-ticking from $458$ to $478$. Past the elbow, the additional keyframes are near-redundant and dilute the generator's fixed compute budget across less informative frames, which delivers noisier supervision to $\pi_{0.5}$. The real-robot side is more sensitive than simulation because fine-motor cues during contact-rich moments are not masked by simulator pose tolerances.

\begin{table}[!htbp]
  \centering
  \caption{\textbf{Keyframe-budget sweep: downstream $\pi_{0.5}$ SR (Mix100).} FVD up-tick and Real-SR drop at $K_{\mathrm{kf}}{=}57$ mark diminishing returns past the elbow.}
  \label{tab:app_kkf_down}
  \scriptsize
  \begin{tabular*}{\linewidth}{@{\extracolsep{\fill}}ccccccccccccc}
    \toprule
    \multirow{2}{*}{$K_{\mathrm{kf}}$} & \multicolumn{5}{c}{LIBERO Sim SR (\%)} & \multicolumn{5}{c}{Franka Real (X/30)} & \multirow{2}{*}{FVD$\downarrow$} \\
    \cmidrule(lr){2-6}\cmidrule(lr){7-11}
    & L-10 & L-Goal & L-Object & L-Spatial & \textbf{Avg} & T1 & T2 & T3 & T4 & \textbf{Avg (\%)} & \\
    \midrule
    9 & 67.5 & 81.7 & 75.1 & 85.7 & 77.5 & 10/30 & 16/30 & 14/30 & 20/30 & 50.0 & 712 \\
    17 & 76.3 & 88.4 & 86.3 & 95.0 & 86.5 & 11/30 & 19/30 & 17/30 & 23/30 & 58.3 & 568 \\
    33 & 87.5 & 93.8 & 92.5 & 96.2 & 92.5 & 11/30 & 21/30 & 18/30 & 27/30 & 64.2 & 489 \\
    \rowcolor{skipblue} \textbf{41 (Ours)} & \textbf{88.8} & \textbf{96.0} & \textbf{94.2} & \textbf{96.4} & \textbf{93.9} & \textbf{12/30} & \textbf{22/30} & \textbf{19/30} & \textbf{28/30} & \textbf{67.5} & \textbf{458} \\
    57 & 88.0 & 95.4 & 93.6 & 96.2 & 93.3 & 11/30 & 20/30 & 18/30 & 27/30 & 63.3 & 478 \\
    \bottomrule
  \end{tabular*}
\end{table}

\paragraph{Recovery diagnostics.}

\textbf{Gap predictor accuracy under distribution shift.}
The gap predictor is trained on ground-truth keyframes but at full-pipeline inference it receives \textsc{SKIP-Generator}-synthesized keyframes, which exhibit appearance drift relative to the originals. Tab.~\ref{tab:app_gap_acc} measures its prediction accuracy on this shifted input. The weighted-average RMSE of $1.548$ frames (MAE $1.010$) corresponds to roughly $0.10$ seconds of timing offset at $16$ FPS. Per-suite trends mirror the end-to-end quality results, namely LIBERO-10 has the largest error ($2.708$-frame RMSE) and also the least stable generated-keyframe appearance, while LIBERO-Spatial has the smallest ($0.779$). This error magnitude has bounded impact on AC-FILM's recursive midpoint interpolation, since the recursion distributes a single gap error across multiple subdivision levels, a partial explanation for why end-to-end SSIM remains above $0.83$ on every suite except LIBERO-10.

\begin{table}[!htbp]
  \centering
  \caption{\textbf{Gap predictor accuracy on \textsc{SKIP-Generator} keyframes.} Frame-level prediction error of the gap predictor when applied to generated (rather than GT) adjacent keyframes, reflecting full-pipeline distribution shift.}
  \label{tab:app_gap_acc}
  \scriptsize
  \begin{tabular*}{\linewidth}{@{\extracolsep{\fill}}lcc}
    \toprule
    Suite & RMSE (frames)$\downarrow$ & MAE (frames)$\downarrow$ \\
    \midrule
    LIBERO-10 & 2.708 & 2.118 \\
    LIBERO-Goal & 1.318 & 0.943 \\
    LIBERO-Object & 1.079 & 0.781 \\
    LIBERO-Spatial & 0.779 & 0.480 \\
    \midrule
    \textbf{Weighted average} & \textbf{1.548} & \textbf{1.010} \\
    \bottomrule
  \end{tabular*}
\end{table}

\begin{figure}[!htbp]
  \centering
  \includegraphics[width=\linewidth]{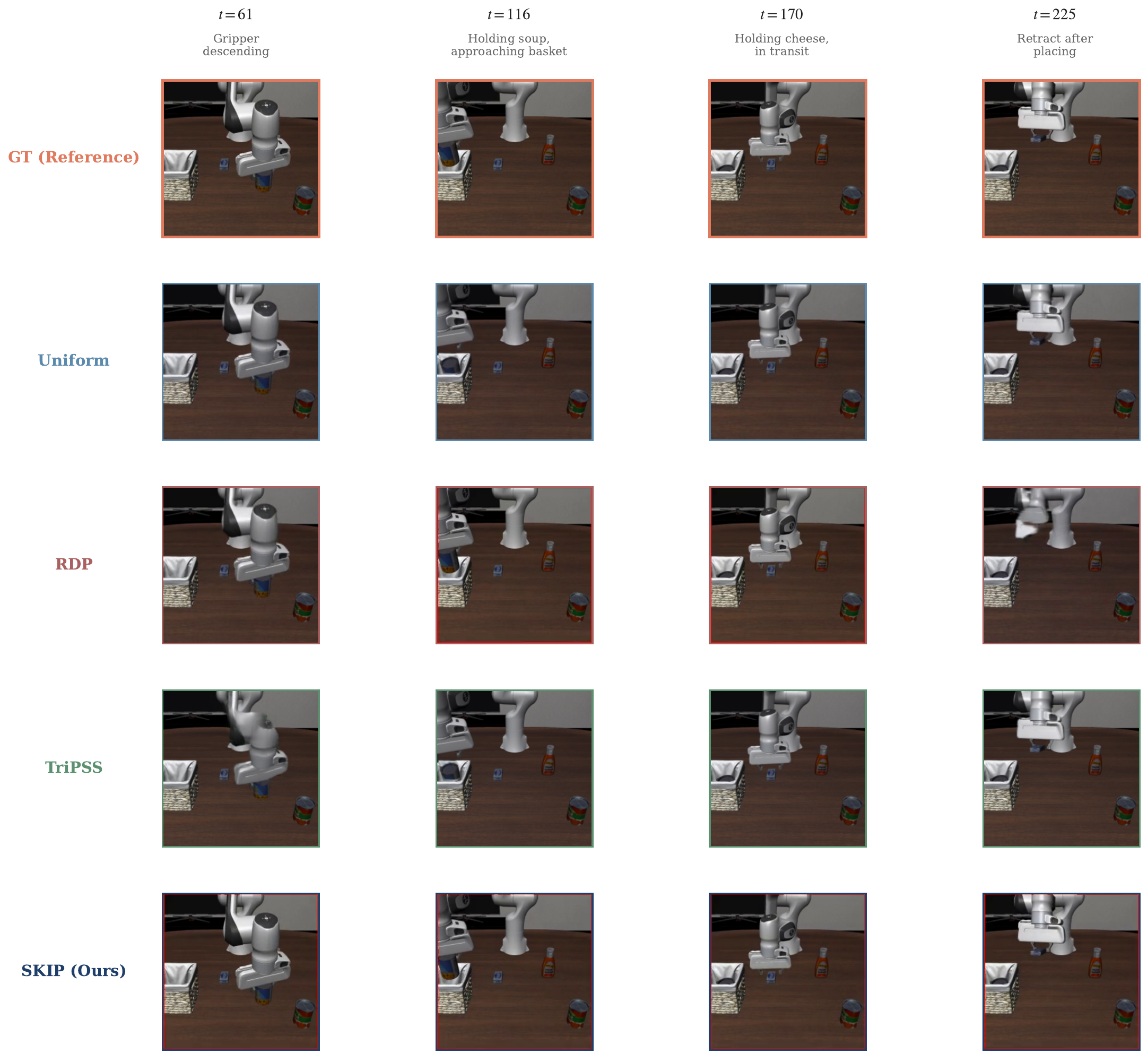}
  \caption{\textbf{Qualitative effect of keyframe placement on dense recovery.} On a long two-object LIBERO trajectory, event-aware keyframes help preserve object-in-gripper states that uniform or geometry-only anchors can miss.}
  \label{fig:app_anchor_qual}
\end{figure}

\textbf{Anchor placement on dense recovery.}
Fig.~\ref{fig:app_anchor_qual} shows a long two-object LIBERO trajectory where event-aware keyframes preserve object-in-gripper states that uniform or geometry-only anchors miss.

\begin{figure}[!htbp]
  \centering
  \includegraphics[width=\linewidth]{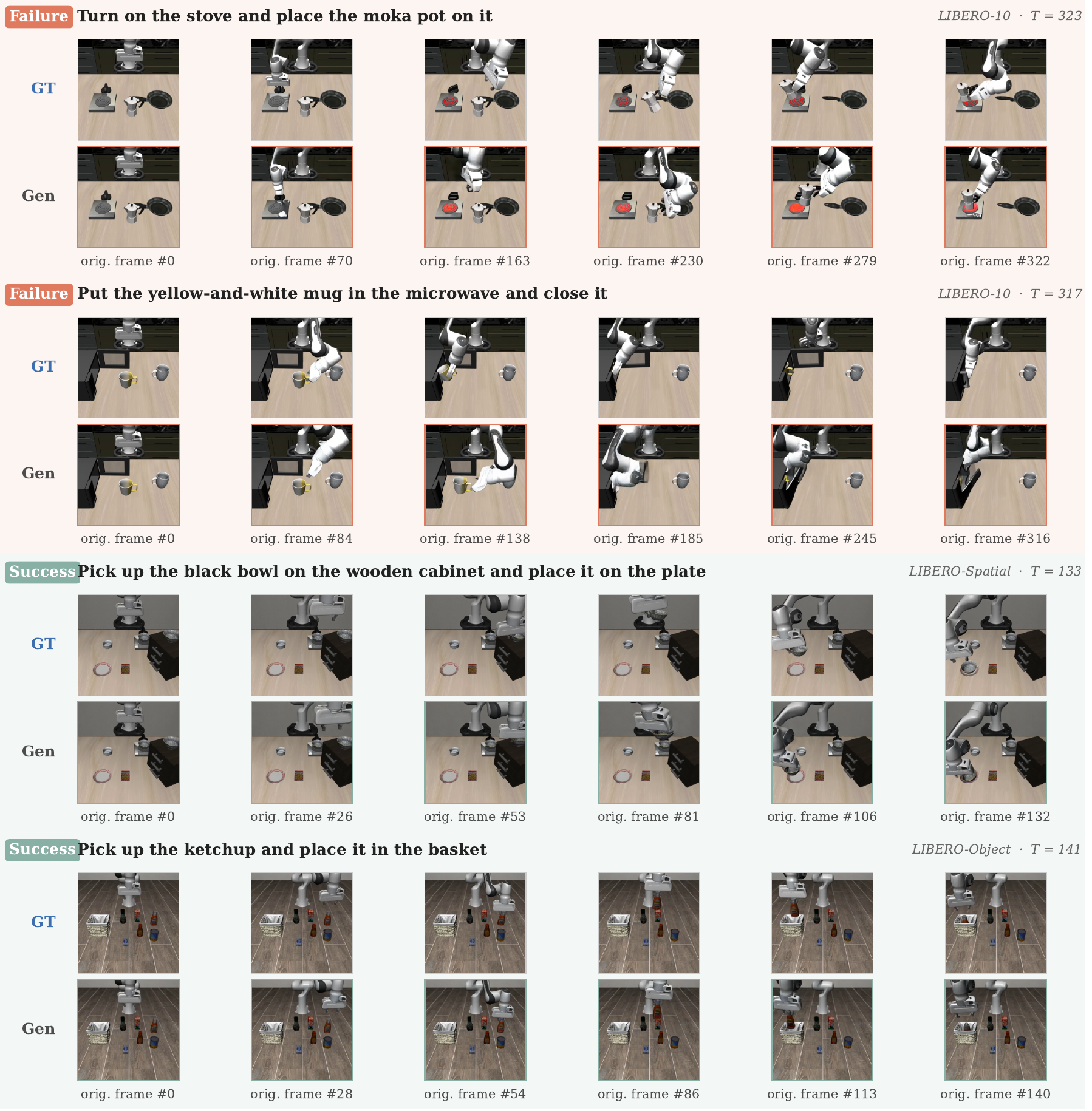}
  \caption{\textbf{Generated dense rollouts versus ground truth.} Sampled frames from \textsc{SKIP}'s reconstructed rollouts (Gen) against ground-truth episodes (GT) on success and failure cases from LIBERO-10, LIBERO-Spatial, and LIBERO-Object. Failures concentrate in compound multi-stage tasks where errors near phase switches accumulate.}
  \label{fig:app_gen_viz}
\end{figure}

\textbf{Generated dense rollouts: failure modes.}
Fig.~\ref{fig:app_gen_viz} shows two recurring failure modes in LIBERO-10's compound tasks. The first is a \emph{phase-switching offset} between subtasks such as opening a stove and placing a moka pot, which causes mid-trajectory object poses to drift from ground truth. The second is \emph{end-stage degradation} on multi-step tasks such as pick-place-and-close-microwave, where the third sub-action accumulates the errors of the first two. Both modes correlate with task length and the number of sub-actions, indicating that the video generator is the residual bottleneck on long compositional rollouts.

\FloatBarrier
\subsection{Policy-data extensions (\S\ref{sec:exp_downstream})}
\label{app:diag_policydata}

\paragraph{Per-condition success rates.}

\textbf{LIBERO simulation per-suite SR.}
As shown in Tab.~\ref{tab:pi05_libero_sr}, SKIP-Mix70 tracks the all-real reference within $1$~pp on every LIBERO suite, and the per-suite difficulty gradient is preserved. LIBERO-10 is the hardest of the four suites under both Real ($90.1\%$) and SKIP-Mix70 ($89.6\%$), while LIBERO-Spatial is the easiest under both ($97.9\%$ Real vs.\ $97.5\%$ SKIP-Mix70). Dense-Mix70 collapses to between $59.4$ and $68.5\%$ and Dense-Mix100 drops further to between $35.5$ and $38.6\%$, with the largest absolute degradation on LIBERO-10 (Real $90.1\%$ vs.\ Dense-Mix100 $35.5\%$, a $54.6$~pp gap). The policy trained on SKIP videos therefore inherits the same task-difficulty profile as the all-real-trained one, while dense-generated rollouts lose information most at long-horizon, high-event-count cells.

\begin{table}[!htbp]
  \centering
  \caption{\textbf{$\pi_{0.5}$ success rate on LIBERO simulation} (\%, per suite). Each condition uses $1503$ training trajectories. Real demos (gray) is the all-real reference, and among the four synthetic-mix conditions the best per column is highlighted.}
  \label{tab:pi05_libero_sr}
  \scriptsize
  \begin{tabular*}{\linewidth}{@{\extracolsep{\fill}}lccccc}
    \toprule
    Training condition & LIBERO-10~$\uparrow$ & LIBERO-Goal~$\uparrow$ & LIBERO-Object~$\uparrow$ & LIBERO-Spatial~$\uparrow$ & \textbf{Avg.}~$\uparrow$ \\
    \midrule
    \rowcolor{black!8}
    Real demos       & 90.1 & 96.5 & 96.3 & 97.9 & 95.2 \\
    \textbf{SKIP-Mix70}  & \cellcolor{skipblue}\textbf{89.6} & \cellcolor{skipblue}\textbf{96.6} & \cellcolor{skipblue}\textbf{96.0} & \cellcolor{skipblue}\textbf{97.5} & \cellcolor{skipblue}\textbf{94.9} \\
    Dense-Mix70  & 59.4 & 67.1 & 68.5 & 64.9 & 65.0 \\
    \textbf{SKIP-Mix100} & 88.8 & 96.0 & 94.2 & 96.4 & 93.9 \\
    Dense-Mix100 & 35.5 & 38.4 & 37.7 & 38.6 & 37.5 \\
    \bottomrule
  \end{tabular*}
\end{table}

\textbf{Franka real-robot per-task SR.}
As shown in Tab.~\ref{tab:realrobot_sr}, SKIP-Mix70 matches Real demos within one success out of $30$ on every real-robot task, and even slightly beats Real on T2 ($26/30$ vs.\ $25/30$). Dense-Mix70 drops to between $5$ and $18$ successes (avg $40.8\%$ vs.\ Real $74.2\%$) and Dense-Mix100 collapses to between $2$ and $11$ successes (avg $25.8\%$). T1 (multi-mug sequential pick-and-place) is hit hardest at $2/30$, matching the long-horizon failure mode that dominates LIBERO-10 in Tab.~\ref{tab:pi05_libero_sr}. The real-robot side amplifies the SKIP-vs-Dense gap relative to simulation because the fine-motor cues that dense $49$-frame chunked generation drifts on are not masked by simulator pose tolerances.

\begin{table}[!htbp]
  \centering
  \caption{\textbf{$\pi_{0.5}$ success rate on Franka Panda real robot.} Each cell reports successes out of $30$ rollouts. Real demos (gray) is the all-real reference, and among the four synthetic-mix conditions the best per column is highlighted.}
  \label{tab:realrobot_sr}
  \scriptsize
  \begin{tabular*}{\linewidth}{@{\extracolsep{\fill}}lccccc}
    \toprule
    Training condition & T1~$\uparrow$ & T2~$\uparrow$ & T3~$\uparrow$ & T4~$\uparrow$ & \textbf{Avg.\ SR (\%)}~$\uparrow$ \\
    \midrule
    \rowcolor{black!8}
    Real demos       & 14/30 & 25/30 & 22/30 & 28/30 & 74.2 \\
    \textbf{SKIP-Mix70}   & \cellcolor{skipblue}\textbf{14/30} & \cellcolor{skipblue}\textbf{26/30} & \cellcolor{skipblue}\textbf{21/30} & 27/30 & \cellcolor{skipblue}\textbf{73.3} \\
    Dense-Mix70  & 5/30 & 16/30 & 10/30 & 18/30 & 40.8 \\
    \textbf{SKIP-Mix100} & 12/30 & 22/30 & 19/30 & \cellcolor{skipblue}\textbf{28/30} & 67.5 \\
    Dense-Mix100 & 2/30 & 10/30 & 8/30 & 11/30 & 25.8 \\
    \midrule
    \multicolumn{6}{l}{\footnotesize T1: \emph{Put both red and blue mugs in basket}. T2: \emph{Put red bowl in drawer and close drawer}.}\\
    \multicolumn{6}{l}{\footnotesize T3: \emph{Put white mug on plate}. T4: \emph{Stack red bowl on green bowl}.}\\
    \bottomrule
  \end{tabular*}
\end{table}

\FloatBarrier
\paragraph{Cross-method downstream comparison.}
To complete the cross-method comparison initiated at the selection-quality level (Tab.~\ref{tab:exp1_main} and Tab.~\ref{tab:app_sel_per_suite}), we run each baseline selection method through the full SKIP pipeline. Its selected keyframes feed \textsc{SKIP-Generator}, then \textsc{SKIP-Reconstructor}, and the resulting synthetic videos train $\pi_{0.5}$ under identical Mix70 and Mix100 protocols to the SKIP-Mix conditions.

\textbf{Aggregate cross-method SR.}
As shown in Tab.~\ref{tab:app_crossmethod_agg}, SKIP-Mix70 outperforms every baseline-Mix70 counterpart by between $7.5$ (vs.\ RDP) and $13.9$~pp (vs.\ TriPSS) in simulation, and by between $15.0$ and $23.3$~pp on the real robot. At Mix100 the gap widens further by an additional $7.2$ to $8.3$~pp in simulation and $2.5$ to $3.3$~pp on the real robot, because the real-data buffer in Mix70 partially compensates for weaker selection. The cross-method ranking by SR (SKIP $>$ RDP $>$ Uniform $>$ TriPSS) follows the OAS ranking from Tab.~\ref{tab:exp1_main}, confirming that selection quality is decisive at the policy-training level.

\begin{table}[!htbp]
  \centering
  \caption{\textbf{Cross-method downstream: aggregate $\pi_{0.5}$ success rate.}}
  \label{tab:app_crossmethod_agg}
  \scriptsize
  \begin{tabular*}{\linewidth}{@{\extracolsep{\fill}}llccc}
    \toprule
    Method & Mix & OAS$\uparrow$ & LIBERO Sim Avg (\%) & Franka Real Avg (\%) \\
    \midrule
    \rowcolor{black!8} Real demos (reference) & n/a & n/a & 95.2 & 74.2 \\
    Uniform-Mix70 & 70 & 0.825 & 84.8 & 54.2 \\
    Uniform-Mix100 & 100 & 0.825 & 75.8 & 45.8 \\
    RDP-Mix70 & 70 & 0.848 & 87.4 & 58.3 \\
    RDP-Mix100 & 100 & 0.848 & 79.2 & 49.2 \\
    TriPSS-Mix70 & 70 & 0.787 & 81.0 & 50.0 \\
    TriPSS-Mix100 & 100 & 0.787 & 71.7 & 41.7 \\
    \rowcolor{skipblue} \textbf{SKIP-Mix70 (Ours)} & 70 & \textbf{0.911} & \textbf{94.9} & \textbf{73.3} \\
    \rowcolor{skipblue} \textbf{SKIP-Mix100 (Ours)} & 100 & \textbf{0.911} & \textbf{93.9} & \textbf{67.5} \\
    \bottomrule
  \end{tabular*}
\end{table}

\textbf{Cross-method per-suite LIBERO breakdown.}
As shown in Tab.~\ref{tab:app_crossmethod_libero}, the SKIP-Mix70-vs-baseline gap is largest on LIBERO-10 (the most event-dense suite in Tab.~\ref{tab:pi05_libero_sr}), where SKIP-Mix70 reaches $89.6\%$ versus $80.0\%$ for RDP-Mix70, $75.0\%$ for Uniform-Mix70, and $67.0\%$ for TriPSS-Mix70. The gap shrinks on LIBERO-Spatial where all four methods exceed $90\%$ at Mix70 because spatial tasks are short-horizon with few gripper events. This per-suite pattern shows that baseline selectors lose ground exactly on the suites that demand event-aware keyframing, which is where \textsc{SKIP-Selector}'s gripper-event protection contributes most.

\begin{table}[!htbp]
  \centering
  \caption{\textbf{Cross-method downstream: per-suite LIBERO Sim SR.} Per-suite breakdown of Tab.~\ref{tab:app_crossmethod_agg}.}
  \label{tab:app_crossmethod_libero}
  \scriptsize
  \begin{tabular*}{\linewidth}{@{\extracolsep{\fill}}lccccc}
    \toprule
    Method (Mix) & LIBERO-10$\uparrow$ & LIBERO-Goal$\uparrow$ & LIBERO-Object$\uparrow$ & LIBERO-Spatial$\uparrow$ & \textbf{Avg}$\uparrow$ \\
    \midrule
    \rowcolor{black!8} Real demos (reference) & 90.1 & 96.5 & 96.3 & 97.9 & 95.2 \\
    Uniform-Mix70 & 75.0 & 86.5 & 84.0 & 93.7 & 84.8 \\
    Uniform-Mix100 & 67.0 & 78.0 & 72.0 & 86.0 & 75.8 \\
    RDP-Mix70 & 80.0 & 87.0 & 88.0 & 94.6 & 87.4 \\
    RDP-Mix100 & 71.0 & 78.0 & 77.5 & 90.3 & 79.2 \\
    TriPSS-Mix70 & 67.0 & 82.0 & 85.0 & 90.0 & 81.0 \\
    TriPSS-Mix100 & 58.0 & 71.0 & 80.0 & 77.8 & 71.7 \\
    \rowcolor{skipblue} \textbf{SKIP-Mix70 (Ours)} & \textbf{89.6} & \textbf{96.6} & \textbf{96.0} & \textbf{97.5} & \textbf{94.9} \\
    \rowcolor{skipblue} \textbf{SKIP-Mix100 (Ours)} & \textbf{88.8} & \textbf{96.0} & \textbf{94.2} & \textbf{96.4} & \textbf{93.9} \\
    \bottomrule
  \end{tabular*}
\end{table}

\textbf{Cross-method per-task real-robot breakdown.}
As shown in Tab.~\ref{tab:app_crossmethod_real}, the per-task pattern mirrors the per-suite finding above. SKIP-Mix70's lead is widest on T1 ($14/30$ vs.\ between $10$ and $13/30$ for baselines) and T4 ($27/30$ vs.\ between $18$ and $22/30$), the two longest-horizon real-robot tasks. T2 and T3 (shorter, fewer events) show smaller gaps, consistent with selection quality mattering less when manipulation events are sparse. At Mix100 the SKIP-Mix advantage on T1 holds up at $12/30$ versus best baseline $11/30$, reproducing the long-horizon resilience seen in LIBERO-10 of Tab.~\ref{tab:app_crossmethod_libero}.

\begin{table}[!htbp]
  \centering
  \caption{\textbf{Cross-method downstream: per-task Franka real-robot SR.} Per-task breakdown of Tab.~\ref{tab:app_crossmethod_agg}, $30$ rollouts per task.}
  \label{tab:app_crossmethod_real}
  \scriptsize
  \begin{tabular*}{\linewidth}{@{\extracolsep{\fill}}lccccc}
    \toprule
    Method (Mix) & T1$\uparrow$ & T2$\uparrow$ & T3$\uparrow$ & T4$\uparrow$ & \textbf{Avg SR (\%)}$\uparrow$ \\
    \midrule
    \rowcolor{black!8} Real demos (reference) & 14/30 & 25/30 & 22/30 & 28/30 & 74.2 \\
    Uniform-Mix70 & 13/30 & 17/30 & 16/30 & 19/30 & 54.2 \\
    Uniform-Mix100 & 10/30 & 15/30 & 13/30 & 17/30 & 45.8 \\
    RDP-Mix70 & 13/30 & 18/30 & 17/30 & 22/30 & 58.3 \\
    RDP-Mix100 & 11/30 & 15/30 & 14/30 & 19/30 & 49.2 \\
    TriPSS-Mix70 & 10/30 & 17/30 & 14/30 & 19/30 & 50.0 \\
    TriPSS-Mix100 & 8/30 & 14/30 & 12/30 & 16/30 & 41.7 \\
    \rowcolor{skipblue} \textbf{SKIP-Mix70 (Ours)} & \textbf{14/30} & \textbf{26/30} & \textbf{21/30} & \textbf{27/30} & \textbf{73.3} \\
    \rowcolor{skipblue} \textbf{SKIP-Mix100 (Ours)} & \textbf{12/30} & \textbf{22/30} & \textbf{19/30} & \textbf{28/30} & \textbf{67.5} \\
    \bottomrule
  \end{tabular*}
\end{table}

\FloatBarrier
\subsection{Ablation extensions (\S\ref{sec:ablation})}
\label{app:diag_ablation}

\textbf{Intermediate selection and recovery metrics.}
Tab.~\ref{tab:ablation_combined} reports the keyframe-quality and dense-recovery metrics that the downstream success rates of Tab.~\ref{tab:ablation_downstream} aggregate, so we can trace which design choice contributes to which SR delta. In panel (a), the proprioceptive stream alone reaches OAS $0.880$ and is the strongest single modality, ahead of V ($0.795$) and S ($0.683$), reflecting that proprio captures contact events that vision struggles to localize. Adding gripper-event post-processing on top of full V+S+P fusion lifts GripCov from $0.842$ to $0.999$ and OAS from $0.863$ to $0.911$, the largest single-step gain in the table. In panel (b), action conditioning lifts dense-recovery PSNR by $+0.730$~dB, SSIM by $+0.008$, and LPIPS by $-0.005$ under identical SKIP-Selector keyframes (paired $t$-test $p<0.01$), confirming that the AC-FILM modulation rather than the keyframe selection is what improves the per-frame fidelity of the recovered rollout.

\begin{table}[!htbp]
  \centering
  \caption{\textbf{Ablations on LIBERO test suites.} Left: modality and gripper-event post-processing ablation for keyframe selection. Right: action conditioning ablation for dense recovery under identical SKIP-Selector keyframes.}
  \label{tab:ablation_combined}
  \scriptsize
  \begin{minipage}{0.55\textwidth}
    \centering
    {\scriptsize\textbf{(a)} Keyframe selection (AC fixed to AC-FILM)}\\[2pt]
    \begin{tabular*}{\linewidth}{@{\extracolsep{\fill}}ccccccc}
      \toprule
      V & S & P & PP & GripCov$\uparrow$ & OAS$\uparrow$ \\
      \midrule
      \cmark & \xmark & \xmark & \xmark & 0.729 & 0.795 \\
      \xmark & \cmark & \xmark & \xmark & 0.396 & 0.683 \\
      \xmark & \xmark & \cmark & \xmark & 0.921 & 0.880 \\
      \cmark & \cmark & \xmark & \xmark & 0.668 & 0.788 \\
      \cmark & \cmark & \cmark & \xmark & 0.842 & 0.863 \\
      \rowcolor{skipblue}\cmark & \cmark & \cmark & \cmark & \textbf{0.999} & \textbf{0.911} \\
      \bottomrule
    \end{tabular*}
  \end{minipage}\hfill
  \begin{minipage}{0.42\textwidth}
    \centering
    {\scriptsize\textbf{(b)} Dense recovery (V/S/P/PP fixed to Ours)}\\[2pt]
    \begin{tabular*}{\linewidth}{@{\extracolsep{\fill}}cccc}
      \toprule
      AC & PSNR$\uparrow$ & SSIM$\uparrow$ & LPIPS$\downarrow$ \\
      \midrule
      \xmark & 30.310 & 0.969 & 0.028 \\
      \rowcolor{skipblue}\cmark & \textbf{31.040} & \textbf{0.977} & \textbf{0.023} \\
      \midrule
      $\Delta$ & $+0.730$ & $+0.008$ & $-0.005$ \\
      \bottomrule
    \end{tabular*}\\[2pt]
    \scriptsize Paired $t$-test $p<0.01$.
  \end{minipage}
\end{table}

\begin{figure}[!htbp]
  \centering
  \includegraphics[width=\linewidth]{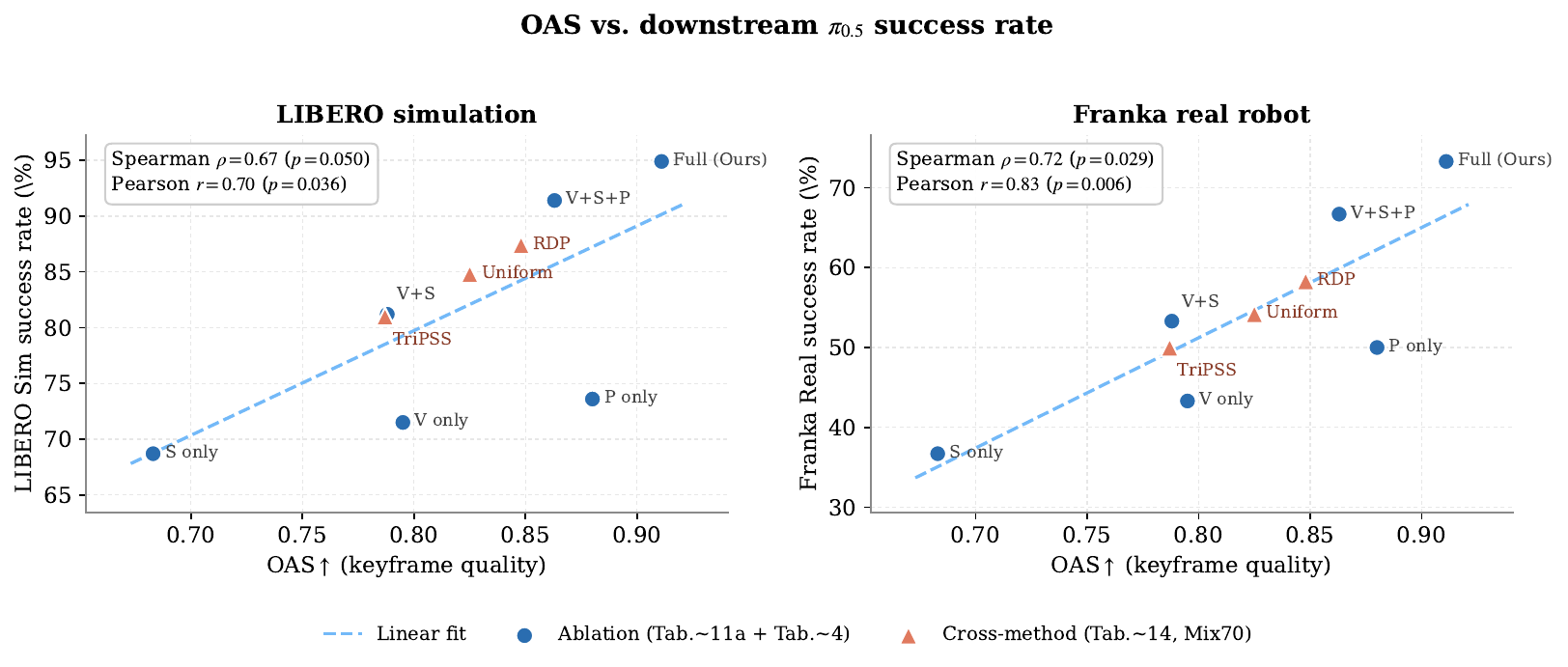}
  \caption{\textbf{OAS tracks downstream $\pi_{0.5}$ success rate.} Circle markers are the six selection-ablation configurations from Tab.~\ref{tab:ablation_combined}(a) paired with their downstream success rates from Tab.~\ref{tab:ablation_downstream} (action conditioning fixed to AC-FILM). Triangle markers are the three cross-method points (Uniform / RDP / TriPSS feeding \textsc{SKIP-Generator} under the Mix70 protocol) from Tab.~\ref{tab:app_crossmethod_agg}. Dashed line is a linear fit across all nine points.}
  \label{fig:oas_sr_corr}
\end{figure}

\textbf{OAS validity check.}
Since OAS is a custom keyframe-quality metric, we verify that it tracks downstream policy performance. We pair the OAS values from Tab.~\ref{tab:ablation_combined}(a) with the corresponding $\pi_{0.5}$ success rates from Tab.~\ref{tab:ablation_downstream} (six selection ablations with action conditioning fixed to AC-FILM), and add the three cross-method points (Uniform, RDP, TriPSS feeding \textsc{SKIP-Generator} under Mix70) from Tab.~\ref{tab:app_crossmethod_agg}. Fig.~\ref{fig:oas_sr_corr} shows the resulting scatter on both LIBERO simulation and the Franka real robot ($N{=}9$ for each panel). The Spearman rank correlations are $\rho=0.67$ on LIBERO sim and $\rho=0.72$ on Franka real, and the Pearson correlations are $r=0.70$ ($p<0.05$) on LIBERO sim and $r=0.83$ ($p<0.01$) on Franka real. The consistent monotonic relationship across two evaluation regimes supports OAS as a meaningful proxy for downstream policy gain.

\textbf{Fusion-weight ablation.}
As shown in Tab.~\ref{tab:fusion_weight}, equal weighting reaches OAS $0.911$ and wins by a small margin over effective-rank-weighted ($0.903$) and fixed-weight ($0.4/0.4/0.2$, $0.900$). Grid search, despite being more sophisticated, scores the lowest at $0.863$ because per-suite weights overfit to the validation split. Looking at the sub-metric breakdown, GripCov stays at $0.999$ across all four strategies because gripper-event post-processing guarantees coverage regardless of fusion weighting. The OAS spread is therefore driven mainly by MEC ($0.888$ for Equal vs.\ $0.800$ for Grid search) and the two semantic-span metrics (MaxSemDist $0.132 \to 0.190$, P95SemDist $0.109 \to 0.157$), confirming that fusion weighting affects multi-event coverage and visual-semantic span but not gripper-event coverage. The narrow $0.011$ OAS gap between equal weighting and the two structured alternatives further indicates that the centering and trace-normalization step already brings the three modalities to comparable scale, so equal weighting is both the simplest and the strongest choice in this regime.

\begin{table}[!htbp]
  \centering
  \caption{\textbf{Fusion-weight ablation on \textsc{SKIP-Selector}.}}
  \label{tab:fusion_weight}
  \scriptsize
  \begin{tabular*}{\linewidth}{@{\extracolsep{\fill}}lccccc}
    \toprule
    Fusion strategy & MEC$\uparrow$ & GripCov$\uparrow$ & MaxSemDist$\downarrow$ & P95SemDist$\downarrow$ & \textbf{OAS}$\uparrow$ \\
    \midrule
    \rowcolor{skipblue}Equal weighting & \textbf{0.888} & \textbf{0.999} & \textbf{0.132} & \textbf{0.109} & \textbf{0.911} \\
    Effective-rank-weighted & 0.875 & 0.999 & 0.145 & 0.117 & 0.903 \\
    Fixed ($0.4/0.4/0.2$) & 0.870 & 0.999 & 0.149 & 0.120 & 0.900 \\
    Grid search & 0.800 & 0.999 & 0.190 & 0.157 & 0.863 \\
    \bottomrule
  \end{tabular*}
\end{table}

\FloatBarrier
\textbf{Generative AI usage disclosure.}
We used large language models (Claude Opus 4.7 and GPT-5.5) for two purposes. For \emph{writing assistance}, they provided grammar correction, style editing, and paraphrasing while polishing an originally Chinese manuscript into English. For \emph{code generation and debugging}, they helped draft and debug the figure-generation and data-analysis scripts. All research ideas, technical content, experimental data, and analyses were produced and verified by the authors.

\end{document}